\documentclass[lettersize,journal]{IEEEtran}
\usepackage{amsmath,amsfonts}
\usepackage{algorithm}
\usepackage{array}
\usepackage{textcomp}
\usepackage{stfloats}
\usepackage{verbatim}
\usepackage{graphicx}
\usepackage{cite}

\usepackage{hyperref}       
\usepackage{breakurl}
\usepackage{placeins}
\usepackage{amssymb}
\usepackage[noend]{algpseudocode}
\usepackage{subcaption}
\usepackage{xcolor}
\usepackage{cancel}

\newcommand{\ch}[1]{#1}
\newcommand{\red}[1]{\textcolor{red}{#1}}
\newcommand{\blue}[1]{\textcolor{blue}{#1}}

\newcommand{\R}{\mathbb{R}}

\renewcommand{\b}{\boldsymbol}

\newcommand{\F}{\mathcal{F}}

\begin{document}

\title{Supervised PCA: A Multiobjective Approach}


\author{Alexander~Ritchie, 
        Laura~Balzano, Daniel~Kessler,
        Chandra~S.~Sripada, and~Clayton~Scott,
\IEEEcompsocitemizethanks{\IEEEcompsocthanksitem A. Ritchie, C. Scott, and L. Balzano are with the Department
of Electrical and Engineering and Computer Science, University of Michigan, Ann Arbor,
MI. E-mail: \{aritch, clayscot, girasole\}@umich.edu\protect
\IEEEcompsocthanksitem C. Scott and D. Kessler are with the Department
of Statistics, University of Michigan, Ann Arbor,
MI. E-mail: \{clayscot,kesslerd\}@umich.edu\protect
\IEEEcompsocthanksitem D. Kessler and C. Sripada are with the Department
of Psychiatry, University of Michigan, Ann Arbor,
MI. E-mail: \{kesslerd,sripada\}@umich.edu\protect
\IEEEcompsocthanksitem C. Sripada is with the Department
of Philosophy, University of Michigan, Ann Arbor,
MI. E-mail: sripada@umich.edu\protect
}
}

\markboth{}%
{}


\maketitle

\begin{abstract}
Methods for supervised principal component analysis (SPCA) aim to incorporate label information into principal component analysis (PCA), so that the extracted features are more useful for a prediction task of interest. Prior work on SPCA has focused primarily on optimizing prediction error, and has neglected the value of maximizing variance explained by the extracted features. We propose a new method for SPCA that addresses both of these objectives jointly, and demonstrate empirically that our approach dominates existing approaches, i.e., outperforms them with respect to both prediction error and variation explained. Our approach accommodates arbitrary supervised learning losses and, through a statistical reformulation, provides a novel low-rank extension of generalized linear models.
\end{abstract}


\section{Introduction}\label{sec:introduction}
\IEEEPARstart{S}{upervised} principal component analysis (SPCA) is, as its name suggests, the problem of learning a low dimensional data representation in the spirit of PCA, while ensuring that the learned representation is also useful for supervised learning tasks. 
SPCA has received considerable interest outside of machine learning \cite{liu2020visualizing, see2020iterative, roberts2006using, chen2008supervised, chen2010pathway, vogelstein2021supervised}, owing to the broad appeal of PCA and a desire to perform supervised dimension reduction. \ch{We view SPCA as a fundamental and important problem, and in this work aim to advance the state of the art in SPCA.}

Toward that end, we introduce a straightforward yet novel approach to SPCA from the perspective of multiobjective optimization. In particular, we propose to solve SPCA by optimizing a criterion that explicitly balances the empirical risk associated to a supervised learning problem with the variance explained by the learned representation.

Compared to prior work on SPCA \cite{bair2006prediction, yu2006supervised, barshan2011supervised, li2016supervised, piironen2018iterative}, our approach has several advantages. First, many prior works are specific to regression or classification, while our approach accommodates arbitrary loss functions. Second, several existing approaches operate in two stages, first learning the representation by one criterion, and subsequently inferring a prediction model by another. These approaches typically use correlation of the learned representation with the response variables as a proxy for the criterion of ultimate interest, e.g., classification accuracy. Third, many existing approaches do not have a means of specifying a trade-off between prediction error (PE) and variation explained (VE), which can lead to poor performance. In our approach, this trade-off is governed by a tuning parameter.

Most importantly, prior research on SPCA has only measured performance in terms of PE, and has not been concerned with whether the learned representation explains substantial variation in the data. Our primary conclusion is that jointly optimizing PE and VE leads to improved generalization. In particular, our approach dominates existing SPCA methods in that it outperforms them in terms of both PE and VE. VE thus serves as a form of regularization for the supervised learning problem, and can also yield more interpretable features.

This paper makes the following contributions.
First, we provide a formulation of SPCA based on multiobjective optimization. Second, we generalize the formulation via a statistical framework, providing a family of SPCA methods similar in spirit to generalized linear models (GLMs). Third, we provide an intuitive maximum likelihood estimation procedure based on manifold optimization. Fourth, we extend the proposed approach to the kernel setting. Finally, we evaluate the proposed approach on real and simulated data, supporting the claims mentioned above.

\vspace{-10pt}
\section{Background and Related Work}
Let $X \in \mathbb{R}^{n \times p}$ be a data matrix whose rows are $p$-dimensional patterns or inputs, and let $Y \in \mathbb{R}^{n \times q}$ be an associated matrix of $q$-dimensional \ch{responses which constitute the target variables of a prediction problem}. The goal of dimensionality reduction (DR) is to find an $r$-dimensional representation of the input data, $r < p$. If $Y$ is used to find this representation, the problem is referred to as supervised dimensionality reduction (SDR). In this section we review PCA, the most common form of DR, as it relates to our contribution. We also review prior work on supervised PCA and other forms of SDR. 

\vspace{-10pt}
\subsection{PCA} \label{pcasection}
PCA was first formulated by Karl Pearson in 1901 \cite{pearson1901liii} and later reinvented by Harold Hotelling \cite{hotelling1933analysis}. Geoemtrically, it can be thought of as the problem of finding an affine subspace of best fit to a collection of points in the squared error sense. As an optimization problem, PCA can be written
\begin{align} 
&\min_{L \in \R^{p \times r}} \ \  \lVert X - XLL' \rVert_F^2  \ \ s.t. \ \  L'L = I_r, \label{eq:pca}
\end{align}
where $X$ is assumed to be centered, $I_r$ is the $r \times r$ identity matrix, and $\|A\|_F$ is the Frobenius norm of a matrix $A$. Projection of $X$ to the subspace spanned by columns of the optimal $L$ gives the best rank-$r$ approximation of $X$ in terms of squared reconstruction error. Equivalently, this projection has the statistical interpretation of capturing the largest possible variance in the data among all rank-$r$ projections. That is, we maximize VE, which is given by
\begin{align}
    \operatorname{VE} &= \frac{\|XL\|_F^2}{\|X\|_F^2} \ \in [0,1]. \label{variation}
\end{align}
Note that this formulation of VE makes sense for any $L$ with orthonormal columns.

The process of performing PCA prior to a regression task is referred to as principal component regression (PCR), a nice discussion of which is given by Jolliffe \cite{jolliffe1982note}. To the authors' knowledge, no such name exists for the analogous approach for classification. This work will refer to that method as principal component classification (PCC).
\vspace{-10pt}

\subsection{Supervised Dimension Reduction}
PCA has enjoyed immense popularity in statistical analysis for the past century or so. It remains a useful tool for dimension reduction (DR) due to its effectiveness, ease of computation and interpretability. However, PCA does not make use of any supervisory information, and therefore DR via PCA may not be useful for subsequent classification or regression tasks. 
This stems from the fact that in most problems of interest, there is a tradeoff between directions that explain variation in $X$, and those that are predictive of $Y$. 
To overcome this limitation, several approaches to SDR have been proposed. We first describe some fundamental SDR methods and highlight their connections to PCA, and then proceed to review existing approaches to SPCA.

Fisher's linear discriminant, or Fisher discriminant analysis (FDA) is arguably the canonical example of supervised dimension reduction in the classification setting. FDA finds a dimension reduced representation of $X$ such that interclass variation is maximized while intraclass variation is minimized. Though it may seem that FDA is generally preferable to PCA for classification, this has been shown not always to be the case, especially when the number of training samples is small \cite{martinez2001pca}. A number of extensions of FDA have been proposed. For example, local Fisher discriminant analysis (LFDA) \cite{sugiyama2007dimensionality} modifies FDA by approximately preserving local distances between same-class points. 

Partial least squares (PLS) regression finds projections of the input data that account for a high amount of variation, but are also highly correlated with projections of the dependent variables. It is somewhat different from other methods presented here, in that both $X$ and $Y$ are projected to a new space to determine their relationship. Without means of specifying the trade-off between correlation and variation, PLS tends to put preference on directions that account for high variation rather than high correlation, causing it to behave similarly to PCR \cite{friedman2001elements}.

Reduced rank regression (RRR) \cite{anderson1951estimating, izenman1975reduced} attempts to minimize regression error under the constraint that the coefficient matrix be low rank. Such models arise in econometrics and other settings where the underlying relationship between predictor and response is believed to be low rank. This model is intimately related to PCA \cite{izenman1975reduced, velu2013multivariate}. Yee and Hastie \cite{yee2003reduced} extend RRR to encompass categorical response variables through what they call reduced rank vector generalized linear models (RRVGLMs). Their work primarily explores the case of reduced rank logistic regression. 

The earliest of the SPCA approaches \cite{bair2006prediction}, which we call Bair's method, is a simple two stage procedure. First, feature selection based on univariate regression coefficients is performed. Second, PCA is performed on the data matrix consisting only of the selected features. This approach may not be optimal, especially in the case where features are jointly predictive but not individually predictive. Furthermore, the method is only applicable to univariate regression and binary classification. On the other hand, this approach has some rigorous theory including a consistency result under an assumption of perfect variable selection with high probability. Recently, the method of iterative supervised principal components (ISPCA) \cite{piironen2018iterative} has extended Bair's method to multiclass classification and reduced computational complexity via an iterative deflationary scheme.

A method herein referred to as Barshan's method \cite{barshan2011supervised} approaches SPCA by means of the Hilbert-Schmidt Independence Criterion (HSIC). In a universal reproducing kernel Hilbert space (RKHS), two random variables are independent if and only if their HSIC is zero. Barshan's method maximizes an empirical measure of the HSIC, which has the form of a trace maximization problem similar to PCA. This method has also been extended to sparse SPCA \cite{sharifzadeh2017sparse}.

A more recent SPCA method, supervised singular value decomposition \cite{li2016supervised} (SSVD), takes a somewhat different approach. They propose an inverse regression model in which $Y$ is a factor in a low rank generative process for $X$. Specifically, the SSVD model has the form
\vspace{-1pt}
\begin{align*}
    X &= UL' + E, \ \ U = YB + F,
\end{align*}
where $E$ and $F$ are error matrices, $U$ is a low-rank score matrix, and $B$ is a coefficient matrix. This method has only been developed for regression.

The approach most similar to our work, and the only SPCA method to model PE directly, is supervised probabilistic principal component analysis \cite{yu2006supervised} (SPPCA). SPPCA extends the probabilistic principal component analysis (PPCA) \cite{tipping1999probabilistic} framework. As with PPCA, the likelihood model of SPPCA allows for statistical testing and Bayesian inference. The method uses an EM algorithm, which can be slow to converge. In addition, this approach places the same amount of emphasis on the dependent and independent variables, and is sensitive to the relative dimensions of $X$ and $Y$. However, SPPCA provides a convenient and straightforward extension to the semi-supervised setting. The relationship of SPPCA to the proposed work is further discussed in $\S$ \ref{sec:sppca}.

Finally, we mention a related line of work \cite{kawano2015sparse, kawano2018sparse, kawano2020sparse}, that takes a regularization approach for adding supervision to the sparse PCA problem \cite{zou2006sparse}.

\ch{The present work is an extension of our preliminary work \cite{ritchie2019supervised}. This preliminary work showed experimentally that the multiobjective approach to SPCA outperforms existing SPCA methods in terms of both PE and VE. These findings are supported by a recent survey of linear supervised dimension reduction methods \cite{xu2021supervised}, which found the approach from our preliminary work to consistently outperform other SPCA methods.}

In the sequel, we extend our preliminary work \cite{ritchie2019supervised} in several ways. First, we motivate our approach from the perspective of multiobjective optimization, which is novel in the SPCA literature, and highlight the interpretation of our criterion as a form of regularized empirical risk minimization. Second, we formulate a statistical model to generalize the optimization formulation.
This allows us to develop a maximum likelihood approach for parameter selection, eliminating the need for a computationally expensive cross-validation (CV) approach, and to draw connections to generalized linear models. Third, we extend our approach to the kernel setting, allowing for nonlinear SPCA. We also include several new experiments to highlight the role of Pareto optimality in SPCA and to show interpretability of the proposed method.

\section{Approach}
We propose an approach to SPCA based on multiobjective optimization that leads to a natural nonstatistical formulation. We then describe a generalization via a statistical model which connects SPCA to generalized linear models. Finally, we extend the method to the kernel setting.
\vspace{-10pt}
\subsection{Notation}
Column vectors will be written as bold lowercase letters. The $i^{th}$ standard basis column vector is written $\b e_i$ and the vector of all ones is written $\b 1$. The $i^{th}$ column of a matrix $A$ is denoted $A_i$ and the entry in the $i^{th}$ row and $j^{th}$ column $A_{ij}$. The transpose of a real valued matrix $A$ is denoted $A'$, while the pseudoinverse is written $A^+$. Bold lowercase letters with positive integer subscripts will refer to realized data samples viewed as column vectors, and will comprise the corresponding data matrices such that $X = [\b x_1 \ \b x_2 \ \dots \ \b x_n]'$ and $Y = [\b y_1 \ \b y_2 \ \dots \ \b y_n]'$. \ch{Throughout this work we assume $\b x_i \in \R^p$ and $\b y_i \in \R^q$, where $\b y_i$ is continuous in the case of regression and one-hot in the case of classification.} To simplify notation, $X$ is assumed to have been centered, meaning the columns have zero mean. $Y$ is assumed to have been centered when its entries are realizations of continuous random variables. Random variables are written as regular font lowercase letters regardless of dimension. The set of positive integers $\{1, 2, \dots, k\}$, is written $[k]$.
\vspace{-5pt}
\subsection{Optimization Formulation} \label{paretosection}
\begin{figure}[htb!]
  \centering
  \includegraphics[width=5 cm]{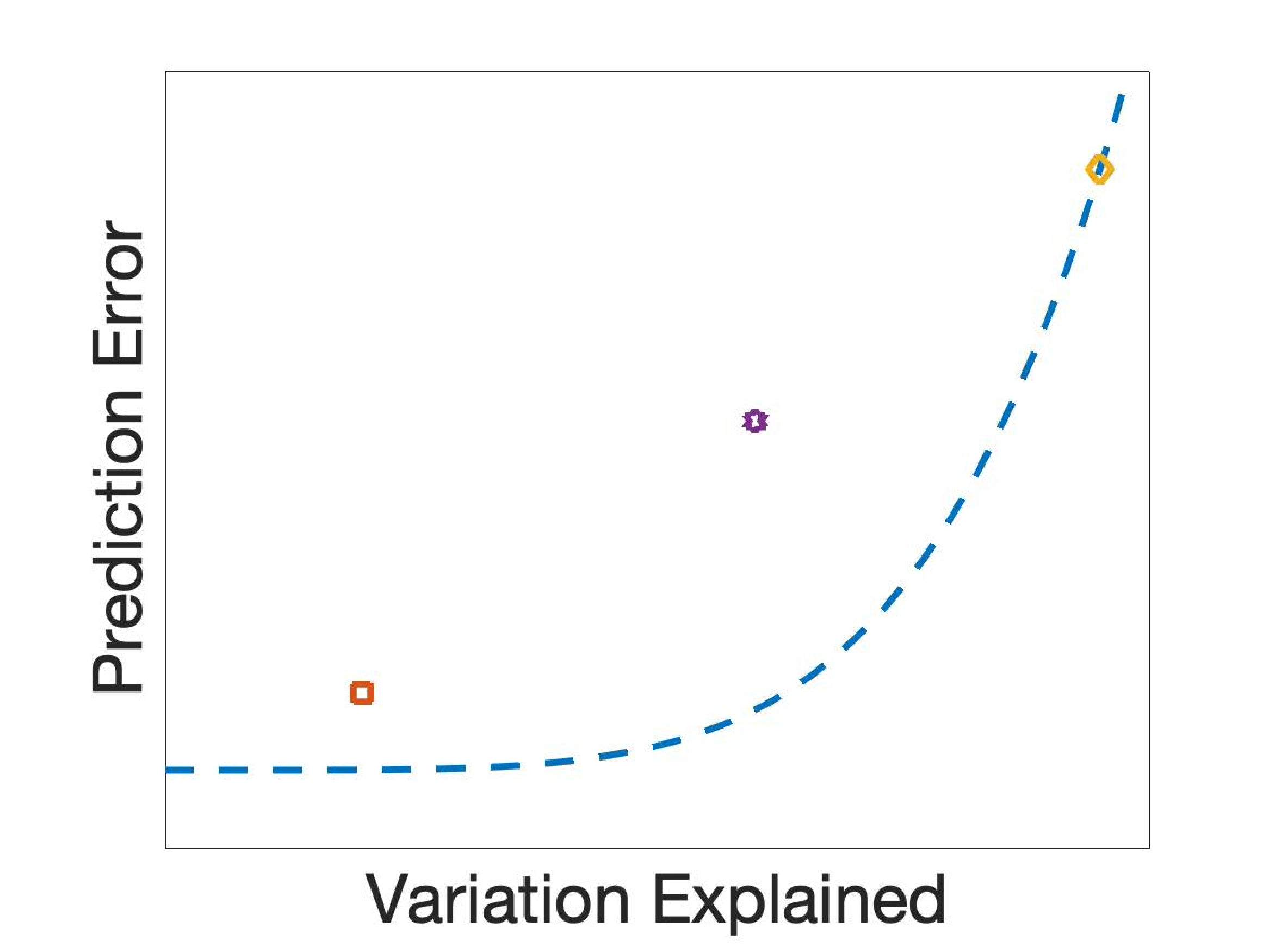}
  \caption{Illustration of Pareto optimality in SPCA. The dashed blue curve represents the Pareto frontier, with the point on this curve representing a single Pareto optimal solution. Solutions above and to the left of the Pareto frontier are suboptimal in both performance measures.}\label{fig:pareto}
\end{figure}

The goal of SPCA is to solve the supervised learning problem while simultaneously performing dimension reduction according to PCA. In other words, any approach to SPCA should learn a feature representation that gives good prediction while explaining as much variation as possible in the data. In general, these two goals are not aligned. Therefore, it is natural to treat SPCA as a multiobjective optimization problem. In multiobjective optimization, Pareto optimal solutions are those for which one objective cannot be improved without sacrificing performance with respect to another. The set of Pareto optimal solutions, called the Pareto frontier, defines a function in the space of performance measures, the epigraph (or hypograph) of which contains all achievable performances on the problem at hand. This concept is illustrated for SPCA in Figure \ref{fig:pareto}.

Our approach to SPCA is to minimize a weighted sum of the PCA objective as given in $\S$ \ref{pcasection} and an empirical risk associated with the prediction problem. \ch{Though straightforward, this regularized empirical risk minimization approach to SPCA did not appear in the literature until recently \cite{ritchie2019supervised}. To the best of our knowledge, there has been no other work, in SPCA or any other context, considering the PCA objective as a regularizer.} This is a direct way to explicitly trade off these two objectives at the expense of adding a tuning parameter, and can be expected to yield a Pareto optimal point that depends on the tuning parameter. The problem is formulated
\begin{align} 
&\min_{L, \beta} \ \ \sum_{i=1}^n g(\boldsymbol{y}_i, \boldsymbol{x}_i, L, \beta) + \lambda \lVert X - XLL' \rVert_F^2 \label{eq:general_spca} \\ 
&s.t. \ \  L'L = I_r, \nonumber
\end{align}
where $g(\cdot)$ is a loss function relating the dimension reduced data to its label, $n$ is the number of observations, $r$ a hyperparameter for subspace dimension, $\lambda > 0$ is a tuning parameter, and the remaining quantities are described in Table \ref{tab:vars}. \ch{We develop SPCA in both the regression and classification settings. While the extension to other losses is straightforward, for concreteness we explicitly consider} the squared error and logistic losses given by
\vspace{-3pt}
\begin{align*}
    g_{\text{LS}}(\boldsymbol{y}, \boldsymbol{x}, L, \beta) &= \|\b y - \beta'L'\b x\|_2^2 \quad \text{and} \\
    g_{\text{LR}}(\boldsymbol{y}, \boldsymbol{x}, L, \beta) &= \log \frac{\exp(\b x'L\beta_{\b y})}{\sum_{j'=1}^q\exp(\b x'L\beta_{j'})},
\end{align*}
respectively, where $\beta_{\b y}$ is the column of $\beta$ corresponding to the class given by one-hot vector $\b y$. These two methods will be referred to as least squares PCA (LSPCA) and logistic regression PCA (LRPCA).
\begin{table}[!htb]
    \centering
    \caption{Description of Key Variables} \label{tab:vars}
    \begin{tabular}{l|l}
        \textbf{VARIABLE}  &\textbf{DESCRIPTION} \\
        \hline \\
        $\underset{n\times p}{X}$ & Data matrix \\
        $\underset{n\times q}{Y}$ & Response variables matrix \\
        $\underset{p \times r}{L}$ &  Basis for the learned subspace\\
        $\underset{n \times r}{XL}$ & Dimension reduced form of $X$ \\
        $\underset{r \times q}{\beta}$ & Learned coefficient matrix \\
        \hline
    \end{tabular}
\end{table}
Note that $L$ is constrained to the Stiefel manifold, i.e., the set of all matrices with orthonormal columns. 

As the solution to \eqref{eq:general_spca} will not in general be given by the SVD, it is necessary to enforce the orthogonality of the columns of $L$ if we hope to recover orthogonal components as in PCA. The primary means of solving such an optimization problem are manifold gradient algorithms which have been thoroughly developed in the literature \cite{edelman1998geometry, absil2009optimization}. Our algorithms will be presented in $\S$ \ref{sec:algs}.
\vspace{-5pt}
\subsection{Statistical Formulation} \label{sec:stat_form}
\ch{The optimization formulation of SPCA given in \ref{eq:general_spca} was introduced in previous work without statistical motivation \cite{ritchie2019supervised}. In this section, we propose a particular statistical model and show that it is a generalization of \eqref{eq:general_spca}. From this perspective, we connect SPCA to a number of well known methods, develop a maximum likelihood approach for setting $\lambda$, and open the door for principled extensions of SPCA (e.g., missing data, complex data, different response models).} The proposed model is as follows
\begin{equation} \label{SPCAmodel}
  x \sim N(0, \sigma_x^2I_p + \alpha LL'), \quad y|x \sim P_{y|x},
\end{equation}
where $\alpha>0$ and $P_{y|x}$ is assumed to be parameterized in terms of $L$, $\beta$, and perhaps additional parameters $\theta$.
Let $\ell_{x}(L,\sigma_x^2,\alpha; \b x_i)$ be the log likelihood function associated with $x$ when $\b x_i$ is observed, and likewise define $\ell_{y|x}(L,\beta,\theta; \b y_i, \b x_i)$. Ignoring additive constants, the negative log likelihood (NLL) can be written
\vspace{-5pt}
\begin{align} 
    G(L&, \beta, \alpha,  \sigma_x^2, \theta; X,Y) \nonumber \\ 
    \triangleq &-\sum_{i=1}^n \ell_{y|x}(L, \beta, \theta; \b y_i, \b x_i) \nonumber - \sum_{i=1}^n\ell_{x}(L,\sigma_x^2,\alpha; \b x_i) \nonumber \\
    = &-\sum_{i=1}^n \ell_{y|x}(L, \beta, \theta; \b y_i, \b x_i) \label{eq:general_stat_obj} \\
    &+ \frac{1}{2\sigma_x^2}\left\|X - \frac{\sqrt{\sigma_x^2 + \alpha} - \sigma_x}{\sqrt{\sigma_x^2 + \alpha}}XLL'\right\|_F^2 \nonumber\\
     &+ \frac{1}{2}\left(n(p-k)\log(\sigma_x^2) + nk \log(\sigma_x^2 + \alpha)\right). \nonumber
\end{align}
The derivation is shown in the appendix.

We are interested in the maximum likelihood estimates (MLEs) of $L$ and $\beta$. The optimization problem is written
\vspace{-2pt}
\begin{align} \label{eq:general_stat}
&\min_{L, \beta, \alpha, \sigma_x^2, \theta} \ \ G(L, \beta, \alpha, \sigma_x^2, \theta; X,Y) \ \ s.t. \ \  L'L = I_r.
\vspace{-2pt}
\end{align}
We consider $\alpha$, $\sigma_x^2$, and $\theta$ to be nuisance parameters, i.e., they are ultimately not of interest but must be accounted for to estimate $L$ and $\beta$. Setting these parameters in practice will be discussed further in $\S$ \ref{sec:param}.

Examining the limiting behavior of $G$ with respect to the nuisance parameters reveals several existing dimension reduction methods to be special cases of the proposed model. As $\alpha \to \infty$, minimizing $G$ with respect to $L$ and $\beta$ can be cast in the form of \eqref{eq:general_spca} where $g = - \ell_{y|x}$. In a similar sense, $G$ approaches the PCA objective as $\sigma_x^2 \to 0$. In the case where $P_{y|x}$ is a generalized linear model (GLM; see $\S$ \ref{sec:glm}), we obtain the RRVGLM corresponding to $\ell_{y|x}$ as $\sigma_x^2 \to \infty$, and the standard GLM corresponding to $\ell_{y|x}$ if $r=p$ (in which case $L$ just represents a change of basis).
\subsubsection{Reinterpreting LSPCA and LRPCA} \label{sec:reinterp} The proposed \\
model accommodates a variety of response models, drawing a parallel to GLMs \cite{nelder1972generalized}. The connection to GLMs is explored further in $\S$ \ref{sec:glm}. For brevity, we explore in detail only the cases where $P_{y|x}$ is Gaussian or categorical.
We now explicitly extend LSPCA and LRPCA using our statistical formulation.

In extending LSPCA we take the response variable to be Gaussian. 
In particular, $y|x \sim N(\beta'L'x, \sigma_y^2 I_q)$ and the NLL, ignoring additive constants, is
\begin{align*}
      G_\text{LS} &= \frac{1}{2\sigma_y^2}\|Y - XL\beta\|_F^2 + nq\log(\sigma_y) - \small{\sum_{i=1}^n\ell_{x}}(L,\sigma_x^2,\alpha; \b x_i).
\end{align*}
Note that with regard to \eqref{eq:general_stat_obj}, we have $\theta = \sigma_y^2$ in this case.

In extending LRPCA we take the response variable to be categorical.
Taking $\b y_i$ to be one-hot vectors encoding class membership, the full NLL, again ignoring additive constants, is
\vspace{-5pt}
\begin{align*}
      G_\text{LR} = &-\sum_{i=1}^n\sum_{j=1}^q\b e_j'\b y_i \log \frac{\exp(\b x_i'L\beta_j)}{\sum_{j'=1}^q\exp(\b x_i'L\beta_{j'})} \\
      &- \sum_{i=1}^n\ell_{x}(L,\sigma_x^2,\alpha; \b x_i).
\end{align*}

Note that with regard to \eqref{eq:general_stat_obj}, there is no $\theta$ in this case as the categorical distribution is completely specified by the class probabilities.

Viewing $G$ as a function of $L$ and $\beta$ with the nuisance parameters held fixed, we may write
\vspace{-5pt}
\small
\begin{equation} \label{eq:general_stat_obj_cv}
     G = \sum_{i=1}^n -\ell_{y|x}(L, \beta, \theta; \b y_i, \b x_i) + \lambda \| X - \gamma XLL'\|_F^2 + c,
     \vspace{-10pt}
\end{equation}
\normalsize
where $c$ is a constant term, $\lambda = \frac{1}{2\sigma_x^2}$ for LRPCA, $\lambda = \frac{\sigma_y^2}{\sigma_x^2}$ for LSPCA, and $\gamma = 1 - (\frac{\sigma_x^2}{\sigma_x^2 + \alpha})^{\frac{1}{2}}$ for both. Moving forward we will use the parameterization in \eqref{eq:general_stat_obj_cv}. In this case, we write the NLL as $G(L, \beta, \lambda, \gamma; X, Y)$. For completeness, we restate the general optimization problem:
\begin{align} 
&\min_{L, \beta, \lambda, \gamma} \ \ G(L, \beta, \lambda, \gamma; X, Y) \label{eq:general_opt} \ \ s.t. \ \  L'L = I_r.
\end{align}
\subsubsection{Connection to Generalized Linear Models} \label{sec:glm}
In the case where $q=1$, we may take $P_{y|x}$ to be a GLM, which is an exponential family model for which there exists a function $h$, called a link function, satisfying $h(\mathbb{E}(Y|X)) = X\beta$. Linear regression and logistic regression are the two most prominent examples. The proposed framework gives rise to a low-rank reformulation of GLMs by adopting the following
modifications:
\vspace{-2pt}
\begin{enumerate}
    \item The link function $h(\mathbb{E}(Y|X)) = X\beta$ is replaced with $\widetilde h(\mathbb{E}(Y|X)) = XL\beta$, which we call the \textit{reduced rank link function}.
    \item The parameters are estimated by optimizing the \textit{joint} log likelihood $\ell_{x,y}$, while parameters for conventional GLMs are estimated by optimizing the \textit{conditional} log likelihood $\ell_{y|x}$.
\end{enumerate}
\vspace{-3pt}
The second point above is critical. In our setting, since $L$ appears in both the marginal likelihood $\ell_x$ and the conditional likelihood $\ell_{y|x}$, optimizing the joint and conditional likelihoods will lead to different estimates of $L$ \emph{and} $\beta$ in general. This differentiates our work from RRVGLMs \cite{yee2003reduced}, of which RRR is a special case, where the conditional likelihood is optimized. As such, the goal of RRVGLMs is to improve out of sample prediction while the focus of this work is SDR.

\subsubsection{Regularization Perspective} \label{sec:reg}
When estimating statistical models in high dimensions, regularization is typically used to avoid overfitting \cite{van2008high, wang2016variable}. 
For instance, ridge regression biases the regression coefficients toward the origin, expressing a degree of belief that the best solution should not have large norm. Alternatively, ridge regression can be viewed as shrinking the effects of the low-variance principal components of $X$ on the regression estimate without ever completely removing them \cite{friedman2001elements}. We can also think of our proposed methods as a form of regularization. For example, the \textit{joint} NLL for LSPCA, restated here for convenience, is
\vspace{-3pt}
\begin{equation*}
     \|Y - XL\beta\|_F^2 + \lambda \|X - \gamma XLL'\|_F^2 + c.
     \vspace{-3pt}
\end{equation*}
The \textit{conditional} NLL consists only of the first term, and its optimum yields the RRR solution. It is clear that minimizing the above with respect to $L$ and $\beta$ will not yield the RRR solution in general. Therefore, optimizing the joint likelihood rather than the conditional likelihood of the proposed models may be thought of as a form of regularization that shrinks the optimal $L$ for RRR toward the PCA solution.
\subsubsection{Connection to SPPCA} \label{sec:sppca}
SPPCA takes a latent variable approach similar to PPCA, extending PPCA to the supervised setting by modeling the conditional distribution of $y$ given the latent variable $z$. Furthermore, SPPCA assumes conditional independence of $y|z$ and $x|z$. The resulting model is
\vspace{-1pt}
\begin{equation*}
  y|z \sim N(V_yz, \sigma_y^2 I_q),  \ x|z \sim N(V_xz, \sigma_x^2 I_p), \  z \sim N(0, \sigma_z^2 I_r),
  \vspace{-1pt}
\end{equation*}
where $V_x$ and $V_y$ are learned parameters modeling $x$ and $y$ as linear functions of $z$, respectively.

The conditional independence assumption may be overly strong, especially when the subspace dimension is misspecified, e.g., the subspace dimension is set too small to capture the full relationship between $y$ and $x$.

Now consider a latent variable model corresponding to LSPCA, where all variables retain their previous definitions:
\begin{equation*}
  y|x \sim N(\beta'L'x, \sigma_y^2 I_q),  \ x|z \sim N(Lz, \sigma_x^2 I_p), \  z \sim N(0, \sigma_z^2 I_r).
  \vspace{-3pt}
\end{equation*}

Empirically we have observed that LSPCA significantly outperforms SPPCA (see $\S$ \ref{sec:exps}). It is clear that the latent variable models differ, though they possess many similarities. We now explore how the differences can explain the proposed method's improved performance. Consider the expectation step in the expectation maximization procedure for SPPCA \cite{yu2006supervised},
\begin{equation*}
    z_i = (\frac{1}{\sigma_x^2}V_x'V_x + \frac{1}{\sigma_y^2}V_y'V_y + I_r)^{-1}(\frac{1}{\sigma_x^2}V_x'x_i + \frac{1}{\sigma_y^2}V_y'y_i),
    \vspace{-5pt}
\end{equation*}
where $z_i \in \mathbb{R}^k$ is the latent representation of the $i^{th}$ data point $(x_i,y_i)$. The above is the MLE of $z|x,y$. At test time, we do not have access to $y$ and so $z$ is taken to be the MLE of $z|x$, which does not depend on $V_y$. This is problematic since $y$ does not depend on $z$ through $V_x$. Therefore, it cannot be assumed that $V_x$ will capture the relationship between $y$ and $z$. Furthermore, if the end goal is to estimate $y$ from $z|x$, it makes sense to directly encode this in the model. This is what LSPCA does. According to the LSPCA model, the MLE of $z|x$ is $z_i = L'x_i$ and $y_i$ only depends on $x_i$ through this quantity. Additionally, this change explicitly shares the parameter $L$ between $P_x$ and the $P_{y|x}$.

To make a direct comparison with SPPCA, we rewrite the LSPCA latent variable model such that $x$ and $y$ are conditioned on the same (reparameterized) latent variable:
\begin{align*}
  y|\widetilde z &\sim N(\beta'\widetilde z, \sigma_y^2 I_q),  \  \
  x|\widetilde z \sim N(L\widetilde z, \sigma_x^2 (I_p - LL')), \\  
  \widetilde z &\sim N(0,(1+\alpha)\sigma_x^2I_r), \\
  \implies x &\sim N(0, \sigma_x^2 (I_p + \alpha LL')).
\end{align*}
Details of the derivation are given in the appendix. The above yields some valuable insight: $y$ and $x$ are conditionally independent given the reparameterization $\widetilde z$ that explicitly assumes $x$ is noiseless in the subspace corresponding to $\widetilde z$. This is a direct result of incorporating the MLE of $z|x$ in the model for $y|x$. While this causes the loss of the conditional independence assumption made by SPPCA, it also causes the MLE of $z|x$ to have a stronger relationship with $y$. Reparameterizing such that $\alpha \gets \sigma_x^2 \alpha$ and integrating out the reparameterized latent variable $\widetilde z$ yields the LSPCA model.
\vspace{-5pt}
\vspace{-1pt}
\subsection{Kernel Supervised Dimension Reduction}
In this section we extend all proposed methods to perform kernel SDR. Further details are provided in the suplementary material.

Kernel PCA (kPCA) \cite{scholkopf1997kernel} is a means of performing non-linear unsupervised dimension reduction by performing PCA in a high-dimensional feature space associated to a symmetric positive definite kernel. Let $k: \R^p \times \R^p \to \R$ be a symmetric positive definite kernel function. Associated to $k$ is a high dimensional feature space $\mathcal{F}$ and mapping $\Phi$ such that $\Phi: \mathbb{R}^p \to \mathcal{F}$. The kernel matrix associated to $k$ is $K_{ij} = k(\b x_i,\b x_j)$, and $k(\b y,\b z) = \langle \Phi(\b y), \Phi(\b z) \rangle_\mathcal{F}$ $\forall \b y, \b z \in \R^p$. Let $X_\Phi$ be the matrix with $n$ rows where each row is the representation in $\mathcal{F}$ of the corresponding row of $X$. Note that kPCA finds the projection of $X_\Phi$ onto its top $r$ principal components, rather than the principal components themselves. Computing the principal components is usually impractical or intractable as $\Phi$ may be unknown and/or $\mathcal{F}$ may be of arbitrarily high or even infinite dimension. Computationally, all that is required is to find the eigenvectors corresponding to the $r$ largest eigenvalues of the centered kernel matrix $\widetilde K = K - \frac{1}{n}\b 1 \b 1'K - \frac{1}{n} K \b 1 \b 1' + \frac{1}{n^2}\b 1 \b 1' K \b 1 \b 1'$. This amounts to solving
\vspace{-2pt}
\begin{align}
   \hat L = &\min_{L} \|\widetilde K - \widetilde KLL'\|_F^2 \label{eq:kPCA} \ \ s.t. \ \  L'L=I_r,
   \vspace{-10pt}
\end{align}
where now $p=n$, and so $L$ is $n \times r$. Noting \eqref{eq:kPCA} has the same form as \eqref{eq:pca}, and that the projection of $X_{\Phi}$ onto its top $r$ principal components is given by $\widetilde K \hat L$, we can kerneliize LSPCA and LRPCA (which we call kLSPCA and kLRPCA, respectively) by simply substituting $\widetilde K$ for $X$ in  \eqref{eq:general_opt}, i.e.,
\vspace{-5pt}
\begin{align} 
&\min_{L, \beta, \lambda, \gamma} \ \ G(L, \beta, \lambda, \gamma; \widetilde K, Y) \ \ s.t. \ \  L'L = I_r.
\vspace{-10pt}
\end{align}
Further details are given in the appendix.
\section{Algorithms} \label{sec:algs}
In this section we present algorithms for solving the optimization problems for LSPCA and LRPCA. We take an alternating optimization approach, breaking the problem into $L$, $\beta$, and nuisance parameter subproblems, wherein one variable will be updated with the others held fixed. The alternating approach allows the algorithm to be extended for response variables modeled by any invertible link function, as long as the inverse link function is differentiable. Let the objective corresponding to the desired method be represented by $G$. Only the linear setting will be described, since in the kernel setting the only difference is the use of the centered kernel matrix $\widetilde K$ directly in place of $X$. We begin by discussing the issue of nuisance parameter selection before discussing the $L$ and $\beta$ subproblems and finally introducing our algorithms.

\vspace{-10pt}
\subsection{Nuisance Parameter Selection} Training of the proposed models requires choosing good nuisance parameter values. We describe two approaches: cross-validation (CV) and maximum likelihood estimation.
\subsubsection{Setting Nuisance Parameters via Cross-Validation}
The proposed models require the estimation of $\alpha$, $\sigma_x^2$, and, in the case of LSPCA, $\sigma_y^2$. When performing CV the model must be trained for each combination of parameter values considered, causing the effective training time to increase rapidly with the number of nuisance parameters. Therefore, it is worth considering if the number of nuisance parameters can be effectively reduced. It was shown in $\S$ \ref{sec:reinterp} that the LSPCA optimization problem can be reparameterized to reduce the number of nuisance parameters to two. We now show that number can be reduced to one for both LSPCA and LRPCA.

For fixed $\sigma_x^2$ and $\alpha$, there is a value of $\lambda$ such that the optimization problems \eqref{eq:general_spca} and \eqref{eq:general_stat} have identical solutions for $L$ and $\beta$. To see this, recall that the problems \eqref{eq:general_stat} and \eqref{eq:general_opt} are equivalent and consider minimizing $\|X - \gamma XLL'\|_F^2 = \|X\|_F^2 - \gamma(2 - \gamma)\operatorname{tr}(X'XLL')$ over $L$. Note that $\gamma \in (0,1)$ does not affect the optimal $L$, just the optimal function value. We can therefore view fixed $\gamma$ (equivalently, fixed $\sigma_x^2$ and $\alpha$) as a scale factor of the PCA term in \eqref{eq:general_spca}, set $\gamma$ to one, and re-scale $\lambda$ accordingly. Therefore, when nuisance parameters are set via CV, minimizing the NLL of the proposed model is equivalent to solving \eqref{eq:general_spca}, which only requires setting $\lambda$.
\subsubsection{Maximum Likelihood Nuisance Parameter Updates} \label{sec:param}
In this section, we present the maximum likelihood updates of the nuisance parameters given $L$ and $\beta$. We will refer to iteratively updating the nuisance parameters in this way as the MLE approach. We show the results here for LSPCA, noting that LRPCA is similar. The derivations are given in the appendix. Given $L$ and $\beta$, the maximum likelihood updates of $\sigma_y^2$, $\sigma_x^2$, and $\alpha$ are
\begin{align}
    \hat\alpha &= \max \left(\frac{1}{nr}\|XL\|_F^2 - \hat\sigma_x^2, 0 \right) \\
    \hat\sigma_x^2 &= \begin{cases}
    \frac{1}{np}\|X\|_F^2 & \hat\alpha = 0 \\
    \frac{1}{n(p-r)}\left(\|X\|_F^2 - \|XL\|_F^2\right) & \hat\alpha > 0 \\
    \end{cases}\\
    \hat\sigma_y^2 &= \frac{1}{nq}\|Y - XL\beta\|_F^2.
\end{align}
The maximum likelihood updates of $\gamma$ and $\lambda$ can then be calculated by substitution.

Since the updates for $\hat \sigma_x^2$ and $\hat \alpha$ depend on each other, practical considerations must be made for a reasonable update procedure. Since $\hat \alpha$ depends on the value of $\hat \sigma_x^2$ while $\hat \sigma_x^2$ depends only on the positivity of $\hat \alpha$, the simplest approach is to set $\hat \sigma_x^2$ first. Since $\alpha=0$ implies $x$ is an isotropic Gaussian random variable, a reasonable assumption for real data is $\alpha>0$. Therefore we suggest initializing $(\hat \sigma_x^2)_0$ from the initial subspace estimate $L_0$ under the assumption that $\hat \alpha>0$. The subsequent update procedure is given in Algorithm \ref{alg:param_updates}. To fully understand Algorithm \ref{alg:param_updates}, it must be viewed in the context of Algorithms \ref{alg:alt} and \ref{alg:sub} where the current values of $L$, $\beta$, and $\alpha$ are passed to Algorithm \ref{alg:param_updates} to update $\alpha$, $\sigma_x^2$ and, for LSPCA, $\sigma_y^2$. In the case of LRPCA, there is no $\hat\sigma_y^2$ to contend with but the updates for $\sigma_x^2$ and $\alpha$ are the same as above.

The biggest benefit of using maximum likelihood updates for the nuisance parameters is the elimination of the computationally burdensome CV procedure. As discussed above, the number of tuning parameters can be reduced to one when using CV. However, this still requires training the model for each parameter value considered. On the other hand, using CV allows for the use of more general criteria for determining the "best" parameter value. For example, one may choose the parameter that yields the best prediction given a certain amount of VE.

\begin{algorithm}
\caption{MLE Nuisance Parameter Updates}\label{alg:param_updates}
 \textbf{Input:} An $n \times p$ data matrix $X$, an $n\times q$ response matrix $Y$, a $p \times r$ orthogonal matrix $L$, an $r \times q$ coefficient matrix $\beta$, a scalar parameter $\gamma$ 
\\
\textbf{Output:} A scalar $\lambda$, a scalar $\gamma$
\begin{algorithmic}[1]
\Procedure{UpdateParams}{$X, Y, L, \beta, \gamma$}
\If{$\gamma>0$} \Comment{Equivalent to $\alpha>0$}
\State $\sigma_x^2 \gets \frac{1}{n(p-r)}\left(\|X\|_F^2 - \|XL\|_F^2\right)$
\Else{}
\State $\sigma_x^2 \gets \frac{1}{np}\|X\|_F^2$
\EndIf{}
\State $\alpha \gets \max(\frac{1}{nr}\|XL\|_F^2 - \sigma_x^2, 0)$
\State $\gamma \gets 1 - (\frac{\sigma_x^2}{\sigma_x^2 + \alpha})^{\frac{1}{2}}$
\If{LSPCA}
\State $\sigma_y^2 \gets \frac{1}{nq}\|Y-XL\beta\|_F^2$
\State $\lambda \gets \frac{\sigma_y^2}{\sigma_x^2}$
\ElsIf{LRPCA}
\State $\lambda \gets \frac{1}{2\sigma_x^2}$ 
\EndIf{} \\
\Return $\gamma, \lambda$
\EndProcedure
\end{algorithmic}
\end{algorithm}

\vspace{-15pt}
\subsection{The \texorpdfstring{$\beta$}{TEXT} Subproblem}
For the squared error and logistic losses, in the linear and kernel settings the $\beta$ subproblem is convex and unconstrained. Therefore, a wide variety of approaches can be utilized. For LSPCA the $\beta$ subproblem is ordinary least squares (OLS) with data matrix $XL$ and response matrix $Y$. Since, $XL \in \mathbb{R}^{n\times r}$, where $r$ is the reduced dimension and likely small, the Cholesky decomposition can be used to efficiently solve the problem with complexity $\mathcal{O}(nr^2 + r^3)$. For LRPCA, the subproblem is logistic regression with data matrix $XL$ and responses $Y$. Common implementations of logistic regression use stochastic gradient or quasi-Newton methods. For our Matlab implementation, the backslash operator for LSPCA and built in logistic regression function for LRPCA were used.
\vspace{-10pt}
\subsection{Grassmannian Constraints for Linear Prediction} \label{linear}
All proposed methods have been presented with the Stiefel manifold constraint $L'L=I_r$. Considering the form of the objectives for LSPCA and LRPCA, the optimal value of the objective functions for a given $L$ only depends on the subspace spanned by the columns of $L$. This can be seen by applying the same rotation to $L$ and $\beta$. In settings such as this, the Grassmann manifold, the set of $r$ dimensional subspaces in $\mathbb{R}^p$, is often used for ease of computation. We will only consider Grassmannian optimization in this work, since this allows projection to the tangent space and geodesic steps can be performed more efficiently. To be clear, even though points on the Grassmannian are subspaces, numerical algorithms require a representation of the subspace to be stored. These representations are taken to be matrices with orthogonal columns that span the subspace.
\vspace{-10pt}
\subsection{The \texorpdfstring{$L$}{TEXT} Subproblem}
Though it is not convex, it is easily shown that the PCA problem on the Grassmannian admits no spurious local optima, i.e., a single critical point is a local minimum and all others are strict saddles or local maxima. Several recent works have studied this setting and shown that gradient descent and several other first order methods almost always avoid strict saddle points \cite{lee2016gradient, lee2019first}. This implies PCA can be solved via Grassmannian gradient descent.

The squared error and logistic losses are convex in $L$, and as a result the Hessian of the $L$ subproblem is the Hessian of the PCA problem on the Grassmannian plus a positive (semi-)definite matrix. This suggests that the $L$ subproblem is well structured in a way that that makes optimization easy. Empirically, we observe that the proposed optimization scheme always converges to a good solution when initialized via PCA.

The $L$ subproblem for LRPCA and LSPCA is solved using manifold conjugate gradient descent (MCG) on the Grassmannian \cite{edelman1998geometry}. We restate the algorithm using our notation in the appendix. In all algorithms we specify a call to $\operatorname{MCG}(G(L), L_0)$, where $G$ is a cost function to be minimized over the Grassmannian and $L_0$ is an initial iterate. We note that, while manifold gradient descent with Armijo line search is guaranteed to converge to a stationary point, no such guarantee exists for manifold conjugate gradient descent. However, it is known that if the algorithm converges to a local minimum, it does so superlinearly \cite{edelman1998geometry}. \ch{It is also worth noting that the per-iteration computational complexity of solving the $L$ subproblem is dominated by calculation of the gradient, which has complexity $\mathcal{O}(p^2r + (p+q)r^2 + pqr)$ for LSPCA and $\mathcal{O}(npq^2r + p^2r)$ for LRPCA. In many problems of interest, it may be assumed that $q$ and $r$ are small, and the per-iteration complexity will be low if $p \ll n$. However, if $p > n$ the $\mathcal{O}(p^2r)$ terms can be reduced to $\mathcal{O}(npr)$ by an alternative factoring.} In any case, we observe excellent performance for the problems considered. We found the implementation of Grassmannian conjugate gradient in Manopt \cite{boumal2014Manopt} to be more efficient than a Matlab only custom implementation. For this reason, we utilize Manopt to solve the $L$ subproblem. 

The necessary (Riemannian) partial derivatives with respect to $L$ are
\vspace{-5pt}
\begin{flalign}
    \operatorname{grad} G_{\text{LS}} &= \nonumber  -(I_p - LL') X'(Y-XL\beta)\beta' & \nonumber \\
    &+\lambda\left(\gamma^2 - \frac{\gamma}{2}\right)(I_p - LL')X'XL & \\
     \operatorname{grad} G_{\text{LR}} &= \nonumber \\ -(I_p-&LL')\sum_{\substack{j\in[q] \\ i\in w_j}} \left( \frac{e^{\b x_i'L\beta_j}\b x_i \sum_{j'= 1}^qe^{\b x_i'L\beta_j}(\beta_{j'}'-\beta_j')}{(\sum_{j'= 1}^qe^{\b x_i'L\beta_j})^2} \right)& \nonumber \\
    + &\lambda \left(\gamma^2 - \frac{\gamma}{2} \right)(I_p - LL')X'XL .&
    \vspace{-10pt}
\end{flalign}
With all the pieces in place, a general alternating algorithm for LSPCA and LRPCA, which extends naturally to the corresponding kernel problems, is given in Algorithm \ref{alg:alt}. Before moving to experiments, we mention an alternative algorithm for LSPCA.
\vspace{-5pt}
\begin{algorithm}
\caption{LSPCA/LRPCA Alternating Algorithm}\label{alg:alt}
\textbf{Input:} An $n \times p$ data matrix $X$, an $n\times q$ response matrix $Y$, a $p \times r$ orthogonal matrix $L_0$ with columns given by the first $r$ principal components of $X$, the reduced dimension $r$, a hyperparameter $\lambda>0$ (if doing CV) \\
\textbf{Output:} The $n \times r$ reduced data matrix $Z^*$, the coefficients $\beta^*$, a $p \times r$ orthogonal matrix $L^*$ such that $Z^* = XL^*$
\begin{algorithmic}[1]
\Procedure{$\operatorname{SPCA}_{\text{alt}}$}{$X, Y, L_0, r, \lambda$}
\item[\quad\quad\quad \ \Comment{Initialize $\gamma$ and $\beta$}]
\State $\gamma \gets 1$
\If{LSPCA}
\State $\beta_0 \gets (XL_0)^+Y$
\ElsIf{LRPCA}
\State $\beta_0 \gets \operatorname{solveLR}(XL_0,Y)$
\EndIf
\State $k \gets 0$
\Repeat
\item[\quad\quad\quad \  \Comment{Optionally, perform nuisance parameter updates}]
\If{MLE}
\State $\gamma,\lambda \gets \operatorname{UpdateParams}(X, Y, L_{k-1} , \beta_{k-1} , \gamma)$
\EndIf
\item[\quad\quad\quad \  \Comment{With $\beta$ fixed, solve for $L$}]
\If{LSPCA}
\State $L_k \gets \operatorname{MCG}(G_{\text{LS}}(L, \beta_{k-1}, \lambda, \gamma; X,Y), L_{k-1})$
\ElsIf{LRPCA}
\State $L_{k} \gets \operatorname{MCG}(G_{\text{LR}}(L, \beta_{k-1}, \lambda, \gamma; X, Y ), L_{k-1})$
\EndIf
\item[\quad\quad\quad \  \Comment{With $L$ fixed, solve for $\beta$}]
\If{LSPCA}
\State $\beta_{k} \gets (XL_{k})^+Y$
\ElsIf{LRPCA}
\State $\beta_{k} \gets \operatorname{solveLR}(XL_{k},Y)$
\EndIf{}
\State $k \gets k+1$
\Until{Convergence}
\State{$Z = XL_k$} \\
\Return $Z, \beta_k, L_k$
\EndProcedure
\end{algorithmic}
\end{algorithm}
\subsection{A Faster Algorithm for LSPCA}
In the case of LSPCA, the optimal $\beta$ given $L$ is the OLS solution. Denote the OLS solution $\beta^*(L) = (XL)^+Y$ as a function of L. We define the objective function
\begin{equation*}
    G_{\text{LS}}^{\text{sub}}(L, \lambda, \gamma; X, Y) \triangleq G_{\text{LS}}(L, \beta^*(L), \lambda, \gamma; X, Y)
\end{equation*}
It is easily observed from the chain rule
\begin{align*}
    \nabla  G_{\text{LS}}^{\text{sub}}(L) &= \frac{\partial G_{LS}(L,\beta^*(L), \lambda, \gamma; X, Y)}{\partial L} \\
    &+ \frac{\partial \beta^*(L)}{\partial L}\underbrace{\frac{\partial G_{LS}(L,\beta^*(L), \lambda, \gamma; X, Y)}{\partial \beta}}_{=0}.
\end{align*}
In words, calculating $\nabla  G_{\text{LS}}^{\text{sub}}$ is the same as calculating the partial derivative of $G_{\text{LS}}$ with respect to $L$ and plugging in $\beta^*(L)$. The same argument applies to the Riemannian gradient. This allows us to eliminate $\beta$ from the optimization problem by simple substitution in the objective and the gradient. Empirically, we observe this approach to be faster than the alternating optimization approach. It is applicable in both the MLE and CV nuisance parameter selection settings. The detailed procedure is given in Algorithm \ref{alg:sub}.
\begin{algorithm}
\caption{LSPCA Substitution Algorithm}\label{alg:sub}
\textbf{Input:} An $n \times p$ data matrix $X$, an $n\times q$ response matrix $Y$, a $p \times r$ orthogonal matrix $L_0$ with columns given by the first $r$ principal components of $X$, the reduced dimension $r$, a hyperparameter $\lambda>0$ (if doing CV) \\
\textbf{Output:} The $n \times r$ reduced data matrix $Z^*$, the coefficients $\beta^*$, a $p \times r$ orthogonal matrix $L^*$ such that $Z^* = XL^*$
\begin{algorithmic}[1]
\Procedure{$\operatorname{LSPCA}_{\text{sub}}$}{$X, Y, L_0, r, \lambda$}
\State $\gamma \gets 1$
\State $k \gets 0$
\Repeat
\item[\quad\quad\quad \  \Comment{Optionally, perform nuisance parameter updates}]
\If{MLE}
\State $\beta \gets (XL_{k-1})^+Y$
\State $\gamma,\lambda \gets \operatorname{UpdateParams}(X, Y, L_{k-1} , \beta , \gamma)$
\EndIf
\item[\quad\quad\quad \  \Comment{Solve for $L$}]
\State $L_k \gets \operatorname{MCG}(G_{\text{LS}}^{\text{sub}}(L, \lambda, \gamma; X, Y), L_{k-1})$
\State $k \gets k+1$
\Until{Convergence}
\State{$Z \gets XL_k$}
\State $\beta \gets Z^+Y$\\
\Return $Z, \beta, L_k$
\EndProcedure
\end{algorithmic}
\end{algorithm}
\section{Experiments} \label{sec:exps}
\ch{To show the utility of our approach for SPCA, we conduct several experiments to compare performance of the proposed methods against existing SPCA methods: Barshan's method, SPPCA, SSVD, and ISPCA (we take ISPCA to have subsumed Bair's method). We also compare against PCR/PCC to demonstrate how each SPCA approach differs from the unsupervised method on which it is based. This also serves as a baseline and sanity check; if any SPCA method consistently performs no better than PCR/PCC in terms of PE, then the utility of that method is unclear. Though the main purpose of these experiments is to compare SPCA methods, we include some general SDR methods for completeness. The general SDR methods include RRR and PLS in the regression setting as well as FDA and LFDA in the classification setting. As Barshan's method is the only competitor that has proposed a kernelized version (kBarshan), we compare the proposed kernel methods against kPCR/kPCC, kBarshan, and kernel LFDA (kLFDA). For our method we give results for the MLE nuisance parameter updates, and nuisance parameter selection via CV. As discussed in $\S$ \ref{sec:stat_form}, $\gamma = 1$ was fixed while CV was performed for $\lambda$ as well as the kernel width, where appropriate. We reserve discussion regarding differences between MLE and CV versions of our methods for $\S$ \ref{sec:mlecv}. }

The datasets used are outlined in Table \ref{tab:datasets}. Most datasets are taken from University of California, Irvine machine learning repository\footnote{\url{https://archive.ics.uci.edu/ml/datasets.php}} (UCI) or the Arizona State feature selection repository\footnote{\url{https://jundongl.github.io/scikit-feature/datasets.html}} (ASU). Where available, dataset specific links are provided in the appendix. We consider datasets in both the $n<p$ and $n>p$ settings. For the Music dataset, we uniformly subsampled $100$ observations for the experiments to obtain a regression dataset in the $n<p$ setting.

In $\S$ \ref{sec:predexps}, results are presented for comparison on the prediction task, as other SPCA works only consider this metric. As such, in $\S$ \ref{sec:predexps} CV is performed to minimize PE. In $\S$ \ref{sec:paretoexps} methods are evaluated on the basis of Pareto optimality.


For each experiment, the best linear and kernel methods (including general SDR methods) are highlighted, while \emph{the best among the SPCA methods in the linear and kernel settings are marked with an asterisk} ($*$). For each experiment $20\%$ of the dataset was uniformly selected at random as an independent test set. For methods that require parameter tuning, not including those using maximum likelihood parameter updates, the remaining $80\%$ of data were then used in a $10$-fold CV procedure. All methods were then trained on the full $80\%$ with the set of parameters leading to smallest CV error, if applicable, before being evaluated on the independent test set. This process, including test set selection, was then repeated $10$ times to produce the results in Table \ref{tab:fixedr_Results}. For all kernel methods, a radial basis function (RBF) kernel was used.
\vspace{-5pt}
\begin{table}[htb!]
\centering
\caption{Description of the datasets used herein. The type field denotes whether the dataset is for regression or classification. In the classification case $q$ is the number of classes, while in the regression case it is the dimension of the response variable.} \label{tab:datasets}
 \begin{tabular}{||c c c c c c||} 
 \hline
 Name & Type & $q$ & $n$ & $p$ & Source\\
 \hline\hline
 Ionosphere & class. & $2$ & $354$ & $34$ & UCI \\ 
 \hline
 Sonar & class. & $2$ & $208$ & $60$ & UCI \\ 
 \hline
 Colon & class. & $2$ & $62$ & $2000$ & ASU\\
 \hline
 Arcene & class. & $2$ & $200$ & $10000$ & ASU\\
 \hline
 Residential & regr. & $2$ & $372$ & $103$ & UCI\\
 \hline
  Music & regr. & $2$ & $100 \ (1059)$ & $116$ & UCI\\
 \hline
 Barshan A & regr. & $1$ & $100$ &  $4$ & \cite{barshan2011supervised}\\
 \hline
 MNIST & class. & $10$ & $60,000$ & $784$ & \cite{lecun1998gradient} \\
 \hline
 FMNIST & class. & $10$ & $60,000$ & $784$ & \cite{xiao2017fashion} \\
 \hline
 HCP & regr. & - & $863$ & $34716$ & \cite{van2008high,sripada2019basic} \\
 \hline
\end{tabular}
\end{table}
\vspace{-10pt}
\subsection{Prediction Performance}\label{sec:predexps}
We first evaluate all the methods for a fixed subspace dimension $r=2$. This process is repeated $10$ times, and results are then averaged to produce the entries in Table \ref{tab:fixedr_Results}. We deliberately choose $r=2$ because this is often the dimension chosen for data visualization. This is meant both to provide some quantitative evaluation of potential visualization and to demonstrate performance in the case where limited memory or other resources make larger representations infeasible. We find our methods achieve better PE than existing SPCA methods in nearly every case and never perform worse than second best among all methods considered. In this setting, both SSVD and SPPCA seem to be heavily biased toward PCR/PCC. We further note that our methods are the only SPCA methods capable of consistently meeting or exceeding the performance of the non-SPCA methods considered. However, in the fixed dimension classification setting it appears LFDA and kLFDA are able to outperform the SPCA methods in several cases, albeit at the expense of substantial VE. 

Next, we repeat the above experiments with the subspace dimension $r\geq2$ chosen via $10$-fold CV. Otherwise, the procedure is identical to that described for the $r=2$ case. The proposed methods perform best among linear SPCA methods in five of six experiments. Additionally, we perform best overall in three of six experiments and are among the top three methods in the remainder. The proposed kernel methods are the best performers in four of six experiments.

\begin{table*}[!htb]
\centering
\captionsetup{size=footnotesize}
\caption{Comparison of mean squared error (regression) or error rate (classification) of competing methods, with standard error. Subspace dimension ($r=2$) was held fixed for results in the first column of each dataset. For results in the second column, subspace dimension was chosen by 10-fold CV. SPCA methods are listed in \textbf{bold}. For each experiment, the best linear method is shown in $\b{\red{red}}$, the best kernel method is shown in $\b{\blue{blue}}$, and the best SPCA methods in the linear and kernel settings are marked with an asterisk ($*$). } \label{tab:fixedr_Results}
\scalebox{0.9}{
\begin{tabular}{|c| c| c| c| c| c| c|}
 \cline{2-7}
 \multicolumn{1}{c}{} & 
 \multicolumn{6}{|c|}{Regression} \\
 \cline{2-7}
  \multicolumn{1}{c|}{} & \multicolumn{2}{|c|}{Residential} &  \multicolumn{2}{|c|}{Barshan A} & \multicolumn{2}{|c|}{Music} \\
  \cline{2-7}
 \multicolumn{1}{c|}{} & $r=2$ & \textbf{CV}& $r=2$ & \textbf{CV}& $r=2$ & \textbf{CV}  \\
   \hline
 PCR & $1.115 \pm 0.462 $ & $0.430 \pm 0.185$ & $0.712 \pm 0.346 $ & $0.401 \pm 0.259$ & $1.930 \pm 0.170 $ & $1.770 \pm 0.164$  \\
 \hline
   PLS & $0.525 \pm 0.218 $ & $0.109 \pm 0.036$ & $\b{\red{0.287 \pm 0.081}} $ & $0.288 \pm 0.081$ & $1.770 \pm 0.151 $ & $\b{\red{1.620 \pm 0.131}}$  \\
  \hline
 RRR & $0.112 \pm 0.091 $ & $0.112 \pm 0.091$ & $0.289 \pm 0.081 $ & $0.289 \pm 0.081$ & $1.633 \pm 0.157 $ & $1.633 \pm 0.157$ \\
 \hline
 \textbf{ISPCA} & $0.380 \pm 0.212 $ & $0.097 \pm 0.050$ & $0.297 \pm 0.094 $ & $0.288 \pm 0.097$ & $1.884 \pm 0.204 $ & $1.751 \pm 0.144$ \\
 \hline
 \textbf{SPPCA} & $1.117 \pm 0.464 $ & $1.097 \pm 0.455$ & $0.323 \pm 0.128 $ & $0.308 \pm 0.120$ & $1.987 \pm 0.167 $ & $1.987 \pm 0.167$ \\
 \hline
 \textbf{Barshan} & $0.684 \pm 0.245 $ & $0.292 \pm 0.085$ & $0.298 \pm 0.094 $ & $*\b{\red{0.287 \pm 0.091}}$ & $1.769 \pm 0.156 $ & $1.691 \pm 0.160$ \\
 \hline
  \textbf{SSVD} & $1.115 \pm 0.459 $ & $0.416 \pm 0.171$ & $0.379 \pm 0.166 $ & $0.398 \pm 0.153$ & $1.931 \pm 0.169 $ & $1.776 \pm 0.169$ \\
 \hline
  \textbf{LSPCA} (CV)& $*\b{\red{0.070 \pm 0.043}} $ &  $*\b{\red{0.060 \pm 0.030}}$ & $0.291 \pm 0.078 $ & $0.294 \pm 0.078$ & $*\b{\red{1.632 \pm 0.156}} $ & $1.667 \pm 0.133$ \\
 \hline
 \textbf{LSPCA} (MLE)& $0.103 \pm 0.112 $ & $0.069 \pm 0.032$ & $*0.289 \pm 0.081 $ & $0.289 \pm 0.081$ & $1.655 \pm 0.142 $ & $*1.642 \pm 0.138$ \\
 \hline
 \hline
 
 kPCR & $1.076 \pm 0.195$ & $0.631 \pm 0.142$ & $0.675 \pm 0.276$ & $0.341 \pm 0.127$ & $2.173 \pm 1.091$ & $2.090 \pm 1.076$ \\
 \hline
 
 \textbf{kBarshan} & $0.899 \pm 0.212$ & $0.761 \pm 0.166$ & $0.276 \pm 0.099$ & $0.269 \pm 0.099$ & $*\b{\blue{2.054 \pm 1.070}}$ & $2.054 \pm 1.077$ \\
 \hline
  \textbf{kLSPCA} (CV)& $*\b{\blue{0.287 \pm 0.121}} $ & $0.138 \pm 0.097$ & $*\b{\blue{0.163 \pm 0.068}}$ & $*\b{\blue{0.162 \pm 0.065}}$ & $2.061 \pm 1.067$ & $*\b{\blue{2.042 \pm 1.069}}$   \\
  \hline
   \textbf{kLSPCA} (MLE)& $0.445 \pm 0.502$ & $*\b{\blue{0.131 \pm 0.096}}$ & $0.223 \pm 0.139$ & $0.284 \pm 0.144$ & $2.114 \pm 1.057$ & $2.057 \pm 1.055$ \\
 \hline

 \multicolumn{1}{c}{} & 
 \multicolumn{6}{|c|}{Classification} \\
 \cline{2-7}
  \multicolumn{1}{c}{} &  \multicolumn{2}{|c|}{Ionosphere} & \multicolumn{2}{|c|}{Colon} & \multicolumn{2}{|c|}{Arcene}\\
  \cline{2-7}
 \multicolumn{1}{c|}{} & $r=2$ & \textbf{CV}& $r=2$ & \textbf{CV}& $r=2$ & \textbf{CV} \\
 
 \hline
 PCC & $0.400 \pm 0.033$ & $0.146 \pm 0.036$ & $0.367 \pm 0.090$ & $0.217 \pm 0.125$ & $0.374 \pm 0.093$ & $0.323 \pm 0.086$ \\
 \hline
   FDA & $0.147 \pm 0.027$ & - & $0.242 \pm 0.107$ & - & $0.228 \pm 0.084$ & - \\
 \hline
 
 LFDA & $0.160 \pm 0.057$ & $0.146 \pm 0.033 $ & $0.225 \pm 0.088$ & $0.208 \pm 0.119 $ & $\b{\red{0.167 \pm 0.049}}$ & $\b{\red{0.169 \pm 0.052}} $ \\
 \hline
 
  RRLR & $0.161 \pm 0.042$ & $0.151 \pm 0.055$ & $0.208 \pm 0.106$ & $\b{\red{0.183 \pm 0.117}}$ & $0.200 \pm 0.112$ & $0.208 \pm 0.078$  \\
 \hline
 \textbf{ISPCA} & $0.163 \pm 0.041$ & $0.134 \pm 0.030$ & $0.217 \pm 0.137$ & $0.258 \pm 0.133$ & $0.313 \pm 0.058$ & $0.269 \pm 0.077$  \\
 \hline
 
 \textbf{SPPCA} & $0.370 \pm 0.047$ & $0.173 \pm 0.042$ & $0.367 \pm 0.090$ & $0.208 \pm 0.132$ &  $0.374 \pm 0.093$ & $0.323 \pm 0.089$ \\
 \hline
 
 \textbf{Barshan} &  $0.146 \pm 0.031$ & $0.144 \pm 0.041$ & $0.258 \pm 0.149$ & $0.258 \pm 0.114$ & $0.344 \pm 0.050$ & $0.349 \pm 0.070$ \\
 \hline
 
  \textbf{LRPCA} (CV) & $0.161 \pm 0.042$ & $*\b{\red{0.127 \pm 0.025}}$ & $*\b{\red{0.192 \pm 0.104}}$ & $*0.200 \pm 0.125$ & $0.200 \pm 0.112$ & $0.223 \pm 0.083$ \\
 \hline

 \textbf{LRPCA} (MLE) & $*\b{\red{0.141 \pm 0.026}}$ & $0.153 \pm 0.046$ & $0.192 \pm 0.125$ & $0.242 \pm 0.144$ & $*0.190 \pm 0.084$ & $*0.195 \pm 0.058$  \\
 \hline
 \hline
 
 kPCC & $0.429 \pm 0.106$ & $0.060 \pm 0.029$ & $0.342 \pm 0.073$ & $0.225 \pm 0.118$ & $0.349 \pm 0.069$ & $0.313 \pm 0.071$ \\
 \hline
 
  kLFDA & $\b{\blue{0.049 \pm 0.03}}$ & $0.057 \pm 0.038$ & $\b{\blue{0.200 \pm 0.131}}$ & $\b{\blue{0.183 \pm 0.110}}$ & $\b{\blue{0.162 \pm 0.050}}$ & $\b{\blue{0.162 \pm 0.040}}$ \\
 \hline
 
 \textbf{kBarshan} & $0.300 \pm 0.045$ & $0.327 \pm 0.124$ & $0.333 \pm 0.162$ & $0.333 \pm 0.162$ & $0.359 \pm 0.048$ & $0.379 \pm 0.081$ \\
 \hline
 
  \textbf{kLRPCA} (CV) & $*0.071 \pm 0.040$ & $*\b{\blue{0.056 \pm 0.030}}$ & $*0.225 \pm 0.111$ & $0.225 \pm 0.111$ & $*0.231 \pm 0.073$ & $*0.215 \pm 0.047$ \\
 \hline
 
  \textbf{kLRPCA} (MLE) & $0.406 \pm 0.115$ & $0.059 \pm 0.030$ & $0.358 \pm 0.088$ & $*0.208 \pm 0.090$ & $0.349 \pm 0.069$ & $0.233 \pm 0.083$ \\
 \hline
 
\end{tabular}
}
\vspace{-10pt}
\end{table*}
\vspace{-10pt}
\subsection{Interpretability}
\begin{figure*}[!htb]
  \centering
  \includegraphics[width=0.65\linewidth]{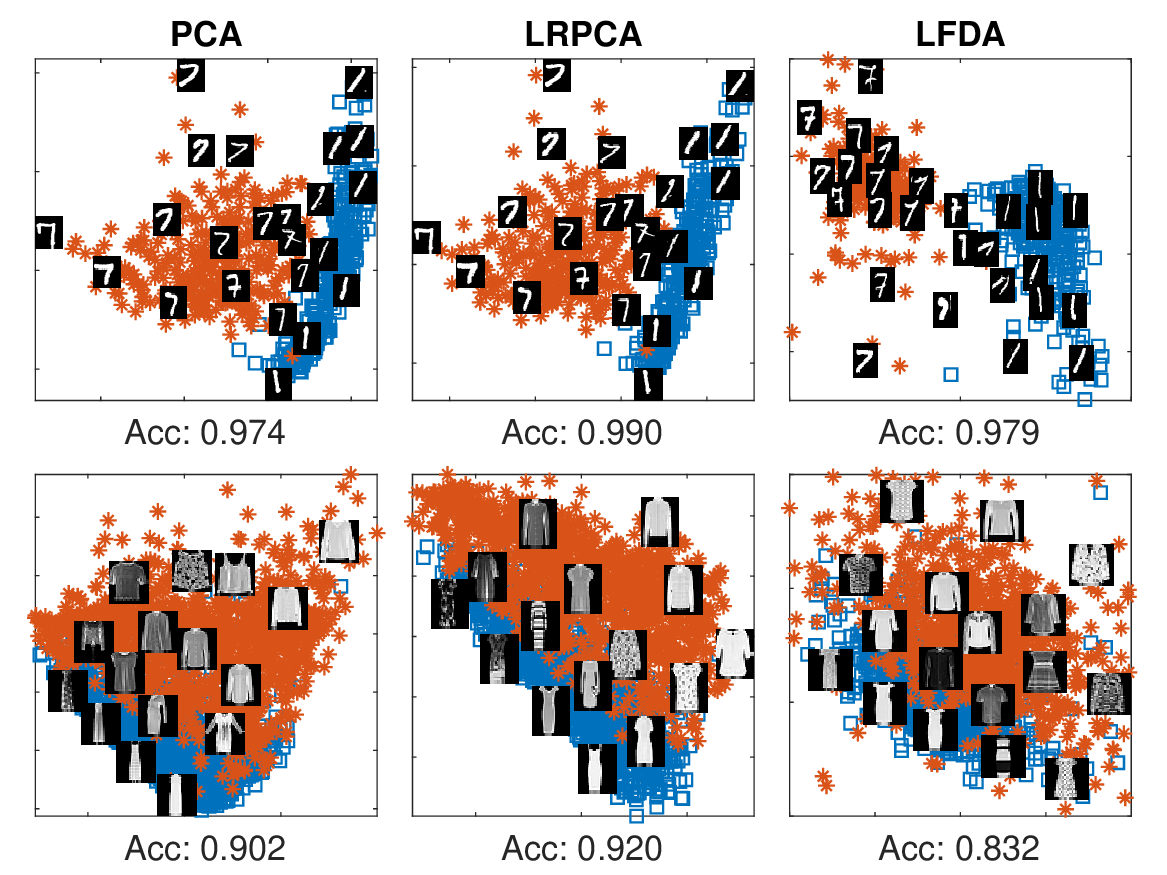}
\caption{Comparison of feature interpretability of PCA, LRPCA, and LFDA on the task of classifying (top) \blue{ones} and \red{sevens} from MNIST and (bottom) \blue{dresses} and \red{shirts} from FMNIST. }
\label{mnistexp}
\vspace{-15pt}
\end{figure*}

In the classification setting LFDA and kLFDA again give the best prediction in several cases, while struggling to represent variation in the data even as higher subspace dimensions are allowed. However, some of our experiments suggest that good prediction without substantial VE can lead to uninterpretable features. Figure \ref{mnistexp} shows test set classification results on MNIST handwritten digits and fashion MNIST (FMNIST) clothing items. We compare embeddings learned by LFDA and LRPCA, since LFDA appears to be the closest competitor in terms of classification accuracy.

For the MNIST experiment, embeddings were learned for the task of binary classification of ones and sevens with subspace dimension $r=2$. The features learned by LSPCA are clearly interpretable. Moving up and to the right along the direction of maximum variation for the ones yields greater clockwise rotation of the vertical section of either digit. Moving up and to the left yields greater length of the horizontal section that distinguishes the digits seven and one. We can also interpret intra-group variation. It appears that ones primarily vary in rotation, tending not to have the horizontal section present in the sevens, while the sevens have substantial variation along both of these features. As expected, points near the boundary between the two classes have small vertical sections, and look like they could be ones or sevens. We note the features learned by LRPCA appear to have the same interpretation as the PCA features, but with improved prediction accuracy. The LFDA embedding has the same property that digits near the boundary have short vertical sections, but there is not the same sense of continuous variation in length of this section. There is no apparent attribute of the digits that changes along the vertical embedding direction. Furthermore, it is difficult to interpret the intra-class variation for either digit. 

For the FMNIST experiment, the experimental setup was identical to the MNIST experiment with the task being binary classification of shirts and dresses. Again the LRPCA and PCA features have similar clear interpretations, with LRPCA having slightly better prediction. In this case, moving up and to the right the length-to-width ratio of the clothing item, an obvious discriminatory feature between shirts and dresses, appears to decrease. Moving up and to the left the brightness of clothing appears to decrease. While not a discriminatory feature, this appears to be a major source of intra-class variation for both shirts and dresses. LFDA appears to learn the length-to-width ratio feature, albeit with substantially worse prediction accuracy. LFDA does not seem to capture intra-class variation.

\begin{figure*}[!htb]
  \centering
  \subcaptionbox{Correlation vs. subpsace dimension.\label{HCPk}}[.37\linewidth]{\includegraphics[width=0.9\linewidth]{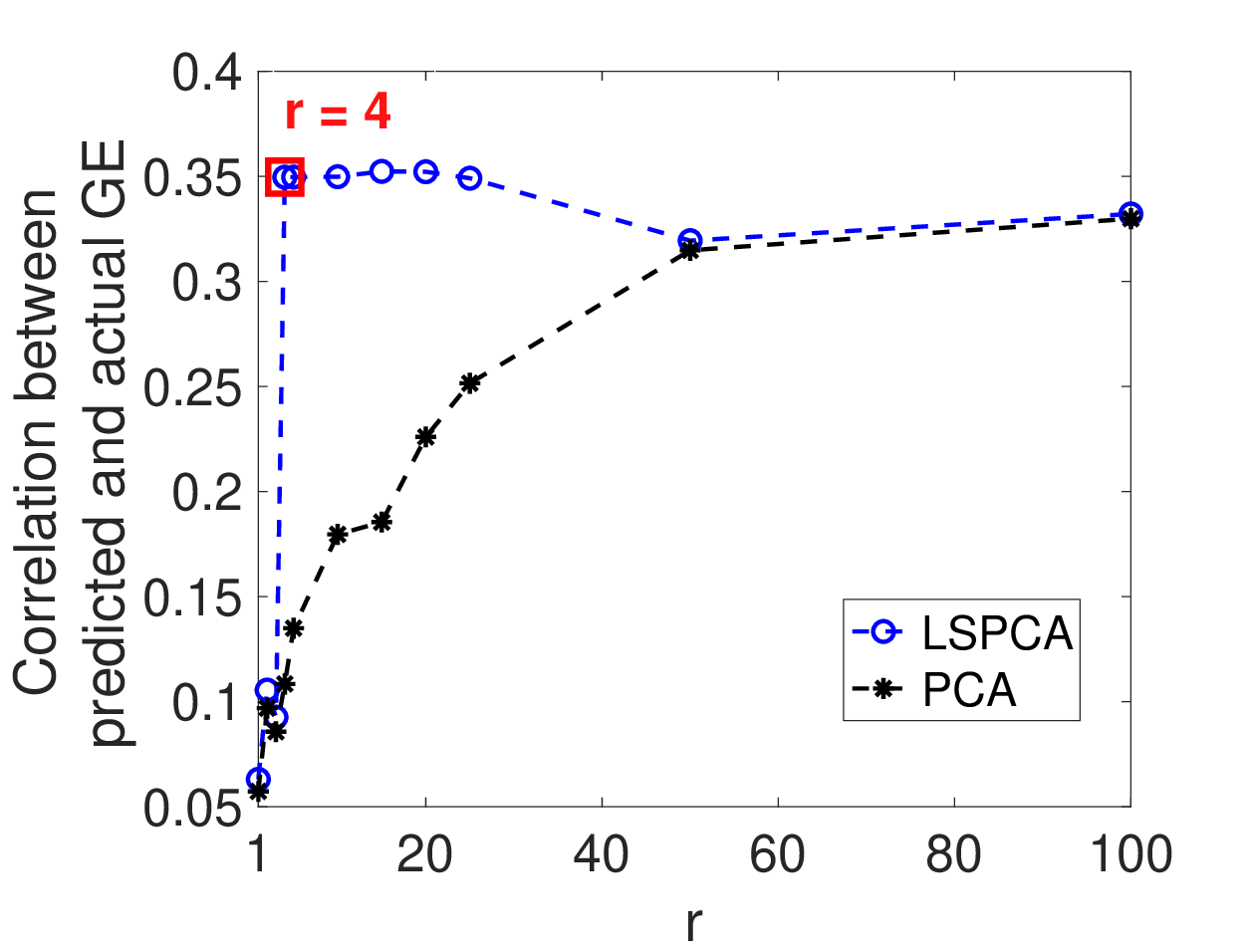}}
  \subcaptionbox{Visualization of components one, three, and four.\label{HCPcomps}}[0.62\linewidth]{\includegraphics[width=0.9\linewidth]{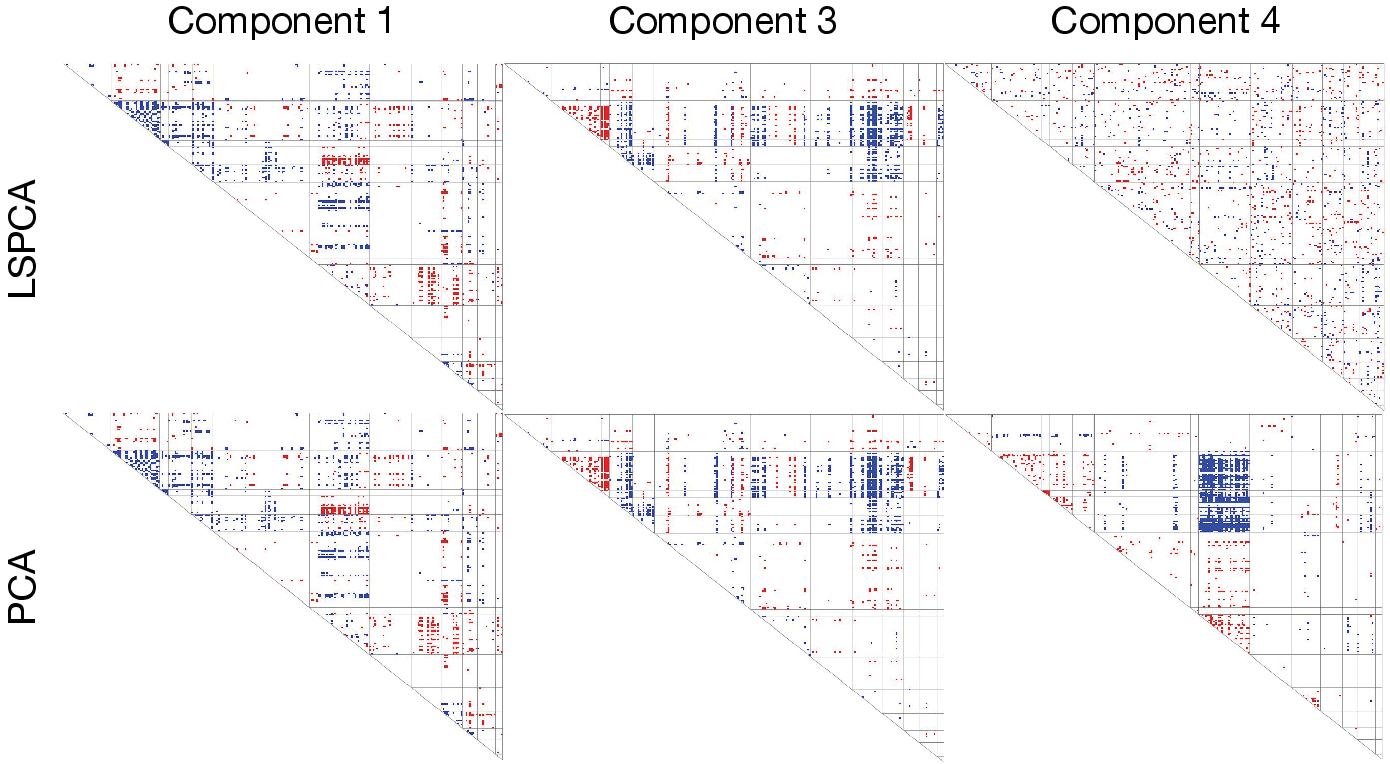}}
  
  \caption{(\ref{HCPk}) Correlation between predicted and actual GE as a function of subspace dimension. (\ref{HCPcomps}) Comparison of components produced by PCA and LSPCA on HCP data, with the components reshaped to reflect  ICN assignments of Power \cite{power2011functional}. \blue{Blue} (\red{red}) denote component entries that are two standard deviations above (below) the component mean. }
  
\label{HCPexp}
\end{figure*}

We consider the application of LSPCA to connectomic data from the Human Connectome Project (HCP) \cite{van2008high}. The data are constructed from functional magnetic resonance imaging of subjects brains, which are time-series, by a number of processing steps, the full details of which are given in \cite{sripada2019basic}. First, for each subject, voxels are collected into a coarse partition consisting of $264$ functional areas, known as the Power parcellation \cite{power2011functional}. Next, voxel-wise behavior is spatially averaged within each functional area, resulting in $264$ time-series from which correlation matrices are constructed. We then construct our data by vectorizing the upper-triangular portions of the subjects' correlation matrices. As in Sripada et al. \cite{sripada2019basic}, our task is to identify patterns of correlated brain activity that predict certain response variables, called phenotypes, associated to each of the subjects, e.g., extroversion, processing speed. Figure \ref{HCPk} shows correlation of actual and predicted General Executive (GE) phenotype \cite{sripada2019basic} on HCP as a function of subspace dimension. The experimental procedure used for this experiment is identical to that described in $\S$ \ref{sec:predexps}. One approach employed in this task is brain basis set (BBS) modeling \cite{sripada2019basic}, which learns a  subspace of small dimension using PCA. LSPCA using $r=4$ is able to achieve equivalent or better predictive performance to BBS with $r=100$, and substantially outperforms BBS with $r=4$. Figure \ref{HCPcomps} shows three of the first four components produced by PCA and LSPCA (PCs and LSPCs, respectively), reorganized according to the intrinsic connectivity network (ICN) assignments of Power \cite{power2011functional}. Components $1-3$ are substantially similar between LSPCA and PCA, but the respective fourth components bear little similarity. Inclusion of the fourth LSPCA component increases average test set correlation of the predicted phenotype from 0.11 to 0.33, while inclusion of the fourth PC increases average correlation from 0.11 to 0.13. The ICN assignments are determined strictly by intra-individual phenomenon, while the PCs and LSPCs are determined by inter-individual variation. It is therefore remarkable that there should be such alignment between PCs and ICN structure (visible in several components depicted in Figure \ref{HCPcomps} where weights concentrate in regions demarcated by overlaid gridlines that reflect ICN structure), a matter which is discussed further by Sripada et al. \cite{sripada2019basic}. However, the fourth LSPC, which is the most predictive component, does not demonstrate substantial ICN structure. This suggests that the bulk of the predictive connectivity for GE is not aligned with the Power ICN, and thus perhaps the structure of the Power ICNs alone are insufficient to fully understand inter-individual differences in GE.
\subsection{Evaluating Pareto Optimality} \label{sec:paretoexps}
\begin{figure*}[!t]
\centering
\begin{subfigure}{.37\textwidth}
  \centering
  \includegraphics[width=\linewidth]{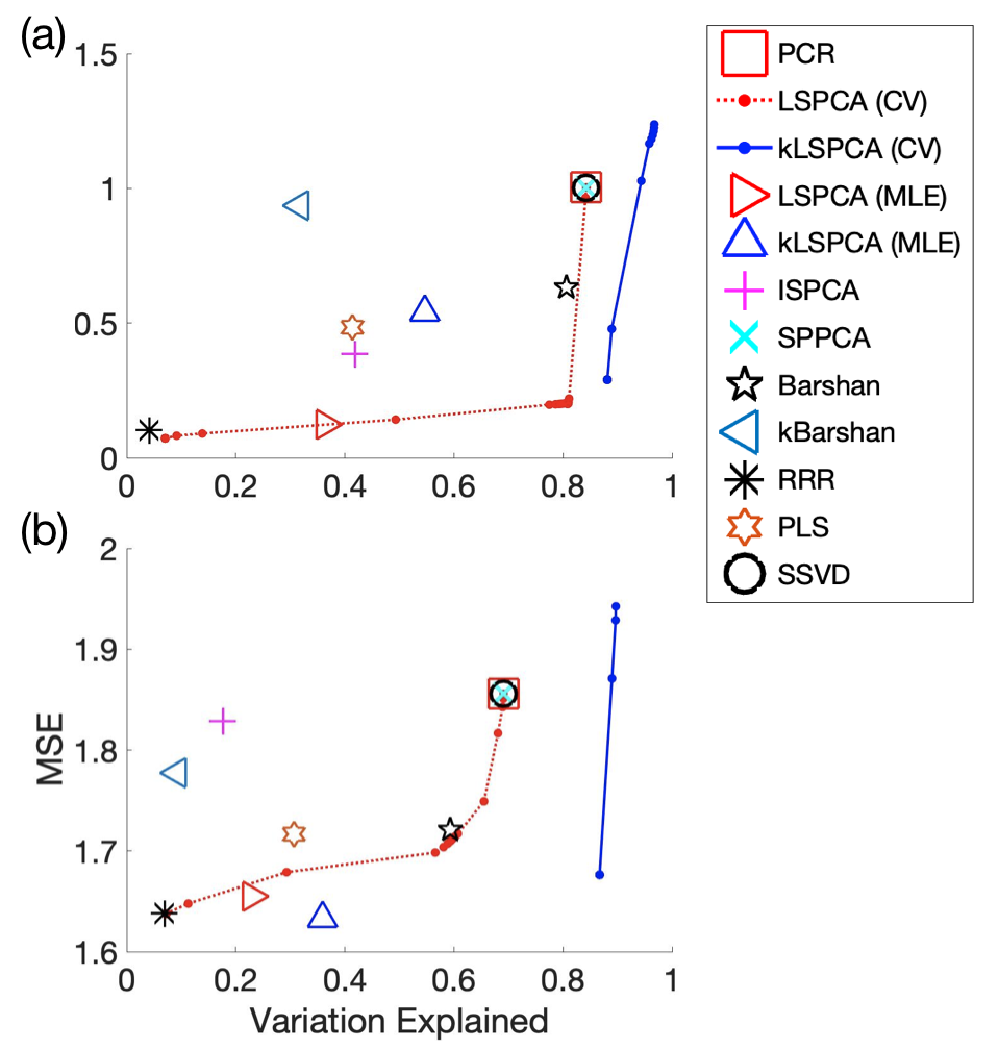}
  \caption{Regression Experiments}
  \label{fig:sub1}
\end{subfigure}%
\begin{subfigure}{.62\textwidth}
  \centering
  \includegraphics[width=\linewidth]{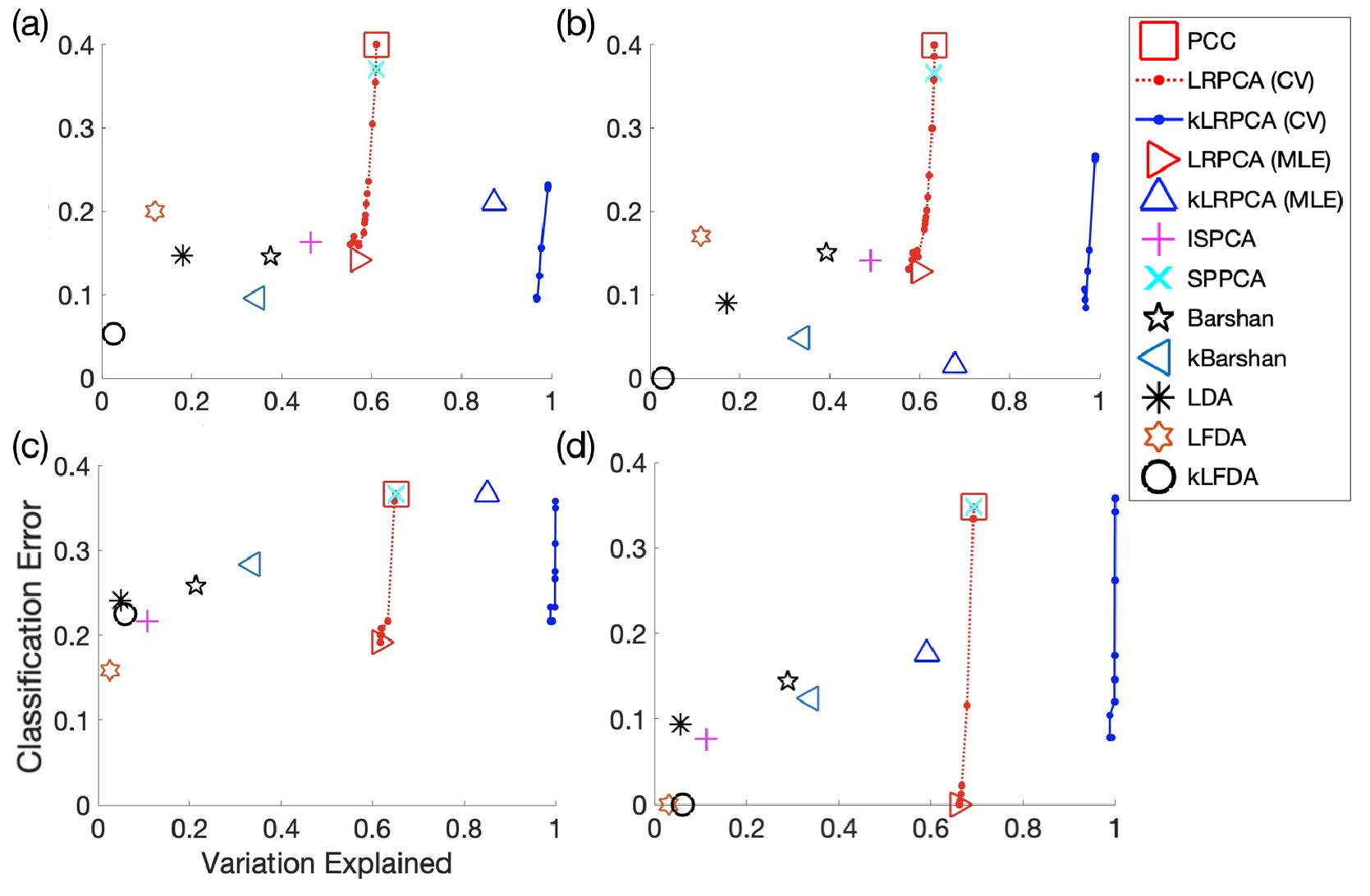}
  \caption{Classification Experiments}
  \label{fig:sub2}
\end{subfigure}
\caption{Comparison of Pareto optimality of competing methods in terms of prediction error and variation explained, with subspace dimension $r=2$. Figure \ref{fig:sub1} shows results for regression datasets (a) residential and (b) music. Figure \ref{fig:sub2} shows results for (a) ionosphere and (c) colon. Additionally Figure \ref{fig:sub2} shows training error for (b) ionosphere and (d) colon.}
\label{paretoexp}
\vspace{-10pt}
\end{figure*}

In this section we compare the proposed approach to competitors through the lens of multiobjective optimization as described in $\S$ \ref{paretosection}. The plots shown in Figure \ref{paretoexp} correspond to the tests in Table \ref{tab:fixedr_Results}, where the dimension $r=2$ is fixed so the comparisons between methods can be direct and meaningful. Solutions that don't generalize to unseen data are of little practical use. We therefore show plots corresponding to training and test data. Figure \ref{fig:sub1} shows plots of VE vs. mean squared error of test data for residential and music datasets. Figure \ref{fig:sub2} shows plots for training and test sets for ionosphere and colon datasets. The curves shown for the CV versions of LSPCA and LRPCA are parameterized by $\lambda$. All plots were generated according to the same procedure described in $\S$ \ref{sec:predexps}.

With regard to the regression experiments the proposed methods dominate all SPCA competitors in the Pareto sense. We remark that the maximum likelihood solution for kLSPCA appears to overfit on the residential dataset, performing worse than CV but still outperforming the kernel version of Barshan's method. In cases where one performance criterion is close, our method always appears to perform significantly better in the other criterion. Moreover, the proposed methods appear able to decrease PE substantially while losing little VE until a point of diminishing returns is reached. After this point, PE can be decreased only marginally for the price of substantial VE. The classification experiments show a similar  pattern. The prominence of this behavior in the training plots suggests that our methods are finding points on or close to the Pareto frontier. Again we see that the MLE approach for kLRPCA appears to overfit. This supports evidence from $\S$ \ref{sec:predexps} that these methods are able to perform well when $r>2$ is allowed, but suffer when subspace dimension is severely restricted.

\renewcommand\thesubfigure{\roman{subfigure}}
\vspace{-10pt}
\subsection{Maximum Likelihood vs. CV}
\label{sec:mlecv}
We find that maximum likelihood nuisance parameter updates often yield prediction performance on par with or better than CV, particularly in higher dimension. However, the $r=2$ experiments show that CV can produce substantially better results for kLRPCA and kLSPCA in the very low dimensional subspace setting. We therefore recommend using CV when $r$ is set very low, as in visualization experiments.
\vspace{-10pt}
\section{Conclusion}
\vspace{-3pt}
We proposed an intuitive, statistically motivated framework for SPCA in various prediction settings. The method generalizes PCA, RRR, and other reduced rank prediction problems, and extends to the kernel setting. We demonstrated that the proposed approach dominates existing SPCA methods and is competitive with other SDR methods in terms of prediction while outperforming them in VE. The proposed maximum likelihood nuisance parameter updates alleviate the need to perform CV, often yielding better prediction, though occasionally sacrificing VE.

The statistical formulation of our approach naturally suggests some directions of future work. For example, a Bayesian approach with sparsifying priors on $L$ and $\beta$ would be of considerable interest. The latent variable interpretation of our method also suggests extensions to standard applications such as missing data and mixture models.

Finally, the use of PCA is ubiquitous in experimental research of the hard sciences, as well as the social sciences. Applying our approach with the proper link functions could yield meaningful insight into important problems. For example, given the prevalence of PCA and ordinal regression tasks in neuroscience \cite{doyle2013multivariate}, biology \cite{dulken2017single}, and other fields, applying our method with the ordered logit response models could be a promising new approach. 
\ifCLASSOPTIONcompsoc
  \section*{Acknowledgments}
\else
  \section*{Acknowledgment}
\fi
This work was supported by ARO W911NF1910027 (AR, LB), AFOSR FA9550-19-1-0026 (AR, LB), NSF IIS-1838179 (AR, LB, CS), NSF CCF-1845076 (LB), NSF CCF-2008074 (CS), NSF DMS-1646108 (DK), a pilot grant U01DA041106 from the Michigan Alzheimer's Disease Center (CSS, AR), a grant R01MH107741 from the Dana Foundation David Mahoney Neuroimaging Program (CSS), and the Michigan Institute for Data Science.

\bibliographystyle{IEEEtran}
\bibliography{IEEEabrv,LSPCA_arxiv.bib}

\appendix

\subsection{Derivation of NLL} \label{likelihood}

In this section we derive, in general terms, the NLL for the proposed model
\begin{equation*}
  x \sim N(0, \sigma_x^2I_p + \alpha LL'), \quad y|x \sim P_{y|x},
\end{equation*}
and put it in functional form of the optimization formulation. From the above, we can write the NLL directly as
\begin{align*} 
    G(L, \beta, \alpha, \sigma_x^2, \theta) \triangleq &-\sum_{i=1}^n \ell_{y|x}(L, \beta, \theta; \b x_i, \b y_i) \\ &- \sum_{i=1}^n\ell_{x}(L,\sigma_x^2,\alpha; \b x_i).
\end{align*}
To supplement what is shown in the main paper, we are interested in finding a simplified form for $\ell_x$. In order to draw a connection to the optimization formulation, we make a few observations. First we can rewrite the covariance matrix of $x$ as
\begin{equation*}
    \sigma_x^2 I_p + \alpha LL' = (\sigma_x I_p + \eta LL')^2,
\end{equation*}
where $\eta = \sqrt{\sigma_x^2 + \alpha} - \sigma_x$. Second, we can write the inverse of the covariance matrix
\begin{align*}
    \left(\sigma_x^2 I_p + \alpha LL'\right)^{-1} &= \left(\sigma_x I_p + \eta LL'\right)^{-2} \\
    &=  \frac{1}{\sigma_x^2}\left( I_p - \frac{\eta}{\sigma_x}LL'\right)^{-2}  \\
    &= \frac{1}{\sigma_x^2}\left( I_p - \frac{\eta}{\sigma_x}L(I_r +  \frac{\eta}{\sigma_x}L'L)^{-1}L'\right)^2 \\
    &= \frac{1}{\sigma_x^2}\left( I_p - \frac{\eta}{\sigma_x}L(\frac{\sigma_x + \eta}{\sigma_x}I_r)^{-1}L'\right)^2 \\
    &= \frac{1}{\sigma_x^2}\left( I_p - \frac{\eta}{\sigma_x + \eta}LL'\right)^2
\end{align*}
where the second step uses the matrix inversion lemma. Third, we simplify the determinant of the covariance matrix
\begin{align*}
    \left|\sigma_x^2 I_p + \alpha LL'\right| &= \sigma_x^{2p}\left| I_p + \frac{\alpha}{\sigma_x^2 } LL'\right| \\
    &= \sigma_x^{2p}\left| I_r + \frac{\alpha}{\sigma_x^2 } L'L\right| \\
    &= \sigma_x^{2p}\left|\frac{\sigma_x^2 + \alpha}{\sigma_x^2 } I_r\right| \\
    &= \sigma_x^{2p}\left(\frac{\sigma_x^2 + \alpha}{\sigma_x^2}\right)^k,
\end{align*}
where the first step makes use of the Weinstein–Aronszajn identity \cite{pozrikidis2014introduction} (sometimes referred to as Sylvester's determinant theorem).

We now rewrite the second term of the NLL omitting additive constants as
\begin{align*}
    -\sum_{i=1}^n\ell_{x}(&L,\sigma_x^2,\alpha; \b x_i) = \frac{1}{2}\operatorname{Tr}\left(X (\sigma_x^2 I_p + \alpha LL')^{-1}X'\right) \\
     & \qquad \quad \quad \ \ \ \quad  + \frac{1}{2} n\log \left|\sigma_x^2 I_p + \alpha LL'\right|  \\
     &= \frac{1}{2\sigma_x^2}\operatorname{Tr}\left(X \left( I_p - \frac{\eta}{\sigma_x + \eta}LL'\right)^2 X'\right) \\
     & \quad + \frac{1}{2}n\log\left(\sigma_x^{2p}\left(\frac{\sigma_x^2 + \alpha}{\sigma_x^2}\right)^k\right)  \\
     &= \frac{1}{2\sigma_x^2}\|X - \frac{\eta}{\sigma_x + \eta}XLL'\|_F^2 \\
     & \quad + \frac{1}{2}\left(n(p-k)\log(\sigma_x^2) + nk \log(\sigma_x^2 + \alpha)\right). \\
\end{align*}
A resubstitution for $\eta$ gives the form shown in the main paper.

\subsection{On SPPCA}
SPPCA takes a latent variable approach similar to PPCA, extending PPCA to the supervised setting by modeling the conditional distribution of $y$ given $z$. Furthermore, SPPCA assumes conditional independence of $y|z$ and $x|z$. The resulting model is
\begin{align*}
  y|z \sim N(W_yz, \sigma_y^2 I_q),  \\ 
  x|z \sim N(W_xz, \sigma_x^2 I_p), \\  
  z \sim N(0, \sigma_z^2 I_r). 
\end{align*}
The conditional independence assumption may be overly strong, especially when the subspace dimension is misspecified, e.g., the subspace dimension is set too small to capture the full relationship between $y$ and $x$.

Now consider a latent variable model for LSPCA, where all variables retain their previous definitions ($L$ still has orthonormal columns):
\begin{equation*}
  y|x \sim N(\beta'L'x, \sigma_y^2 I_q),  \ x|z \sim N(Lz, \sigma_x^2 I_p), \  z \sim N(0, \sigma_z^2 I_r). 
\end{equation*}
 Conditioning $y$ on $x$ alleviates the issue caused by the conditional independence assumption of SPPCA. Forming the joint distribution of $x$ and $y$, ignoring log terms and additive constants, and integrating out the latent variable yields
    \begin{align*}
        f_{x,y}(x,y) = &f_{y|x}(y|x) \int_{-\infty}^{\infty} f_{x|z}(x)f_z(z) dz \\
        \propto &\exp(-\frac{1}{2\sigma_y^2}\|y - \beta'L'x\|_2^2 \\ &-\frac{\sigma_z^2}{2\sigma_x^2(\sigma_x^2 + \sigma_z^2))}x'(\frac{\sigma_x^2 + \sigma_z^2}{\sigma_z^2}I_p - LL')x)
    \end{align*}
where the $f_{(\cdot)}$ are the corresponding density functions, and logarithmic terms are ignored. If we write $\sigma_z^2 = \alpha \sigma_x^2$ with $\alpha = 2\eta\sigma_x + \eta^2$ (implying $\eta = \sqrt{\sigma_x^2 + \alpha} - \sigma_x$, as before) the negative log likelihood evaluated on data matrices $X$ and $Y$ reduces to
\begin{align}
      -G_{\text{LS}} &\propto \|Y - XL\beta\|_F^2 \label{eq:general_nll} + \frac{\sigma_y^2}{\sigma_x^2} \|X(I_p - \frac{\eta}{\sigma_x + \eta}LL')\|_F^2, \nonumber
\end{align}
which matches the form of LSPCA.

Crucially, how should the conditioning of $y$ on $x$ rather than on $z$ be interpreted? In the suggested model
\begin{align*}
    x &= Lz + \zeta_x  \implies y = \beta'(z + L'\zeta_x) + \zeta_y,
\end{align*}
where $\zeta_x \sim N(0, \sigma_x^2I_p)$ and $\zeta_y \sim N(0, \sigma_y^2I_q)$. 
First note that $L'\zeta_x \sim N(0, \sigma_x^2I_r)$, i.e., it is isotropic Gaussian noise in the latent subspace. With the substitution $\sigma_z^2 = \alpha \sigma_x^2$ the model becomes
\begin{align*}
    x &= L(z + \frac{z'}{\sqrt{\alpha}}) + (I_p-LL')\zeta_x\\
    &= L\widetilde z + (I_p-LL')\zeta_x,\\
    y &= \beta'(z + \frac{z'}{\sqrt{\alpha}}) + \underbrace{\beta'L'(I_p-LL')\zeta_x}_{=0} + \zeta_y \\
    &= \beta'\widetilde z + \zeta_y
\end{align*}
where $z,z' \overset{\text{i.i.d.}}{\sim} N(0, \alpha\sigma_x^2 I_r)$ and $\widetilde z \sim N(0, (1+\alpha)\sigma_x^2 I_r)$. We can now write the conditional distributions of $y$ and $x$ on $z$
\begin{align*}
  y|\widetilde z &\sim N(\beta'\widetilde z, \sigma_y^2 I_q),  \\
  x|\widetilde z &\sim N(L\widetilde z, \sigma_x^2 (I_p - LL')), \\  
  \widetilde z &\sim N(0,(1+\alpha)\sigma_x^2I_r), \\
  \implies x &\sim N(0, \sigma_x^2 (I_p + \alpha LL')).
\end{align*}
Reparameterizing such that $\alpha \gets \sigma_x^2 \alpha$ and integrating out the reparameterized latent variable $\widetilde z$ yields the LSPCA model.

\subsection{MLEs of the Nuisance Parameters} \label{sec:param_supp}
In this section we derive the MLE updates of the nuisance parameters. For LRPCA, the nuisance parameters are $\sigma_x^2$ and $\alpha$, while LSPCA adds an additional nuisance parameter $\sigma_y^2$. The partial derivatives w.r.t. $\sigma_x^2$ and $\alpha$ will be the same for $G_\text{LS}$ and $G_\text{LR}$, implying the MLEs $\sigma_x^2$ and $\alpha$ will also be the same for both problems. Therefore, we derive the MLEs for LSPCA only.

Taking the partial derivative of $G_\text{LS}$ with respect to $\sigma_y^2$ yields
\begin{equation*}
    \frac{\partial G_\text{LS}}{\partial \sigma_y^2} = \frac{1}{\sigma_y^2}\left( -\frac{1}{\sigma_y^2}\|Y-XL\beta\|_F^2 + nq\right)
\end{equation*}
which has a single zero at $\sigma_y^2 = \frac{1}{nq}\|Y-XL\beta\|_F^2$. The second partial derivative is positive at this point, making this the unique minimal $\sigma_y^2 $.

Looking at the partial derivative with respect to $\alpha$ yields
\begin{equation*}
    \frac{\partial G_\text{LS}}{\partial \alpha} = -\frac{1}{(\sigma_x^2 + \alpha)^2}\|XL\|_F^2 + \frac{nr}{\sigma_x^2 + \alpha}
\end{equation*}
which has a single zero at $\alpha = \frac{1}{nr}\|XL\|_F^2 - \sigma_x^2$. Furthermore, $G_\text{LS}$ is strictly increasing for $\alpha > \frac{1}{nr}\|XL\|_F^2 - \sigma_x^2$ and strictly decreasing for $\alpha < \frac{1}{nr}\|XL\|_F^2 - \sigma_x^2$, making this critical point a minimizer. Given the nonnegativity constraint on $\alpha$, note this also implies that if $\frac{1}{nr}\|XL\|_F^2 - \sigma_x^2 < 0$, then the minimizer is $\alpha=0$.

For the partial derivative with respect to $\sigma_x^2$ we have
\begin{align*}
    \frac{\partial G_\text{LS}}{\partial \sigma_x^2} = &-\frac{1}{\sigma_x^4}\|X\|_F^2 + \frac{\alpha(2\sigma_x^2 + \alpha)}{\sigma_x^4(\sigma_x^2 + \alpha^2)}\|XL\|_F^2 \\
    &+ \frac{n(p-r)}{\sigma_x^2} + \frac{nr}{\sigma_x^2 + \alpha}.
\end{align*}
Evaluating the above at the optimal $\alpha$, we find that if $\alpha=0$, $\sigma_x^2 = \frac{1}{np}\|X\|_F^2$, which is exactly what is expected from the model. On the other hand, if $\alpha > 0$ we can note the following. The partial derivative is zero when
\begin{align*}
    0 = &-(\sigma_x^2 + \alpha)^2\|X\|_F^2 + \left(\alpha\sigma_x^2 + \alpha(\sigma_x^2 + \alpha)\right)\|XL\|_F^2 \\
    &+ (p-r)\sigma_x^2(\sigma_x^2 + \alpha)^2 + nr\sigma_x^4(\sigma_x^2 + \alpha).
\end{align*}
Plugging in $\alpha = \frac{1}{nr}\|XL\|_F^2 - \sigma_x^2$ yields the optimality condition
\begin{equation*}
    0 = \sigma_x^2 \frac{p-r}{nr^2}\|XL\|_F^4 + \frac{1}{n^2r^2}\|XL\|_F^4(\|XL\|_F^2-\|X\|_F^2),
\end{equation*}
which implies $\sigma_x^2 = \frac{1}{n(p-r)}(\|X\|_F^2 - \|XL\|_F^2)$. In either case $G_\text{LS}$ is strictly decreasing as $\sigma_x^2$ approaches the critical point from the left, and strictly increasing as $\sigma_x^2 \to \infty$ from the right of the critical point, making the corresponding critical points minimizers.

In summary, given $L$ and $\beta$, the maximum likelihood estimates of $\sigma_y^2$, $\sigma_x^2$, and $\alpha$ are
\begin{align}
    \hat\alpha &= \max(\frac{1}{nr}\|XL\|_F^2 - \hat\sigma_x^2, 0) \\
    \hat\sigma_x^2 &= \begin{cases}
    \frac{1}{np}\|X\|_F^2 & \hat\alpha = 0 \\
    \frac{1}{n(p-r)}\left(\|X\|_F^2 - \|XL\|_F^2\right) & \hat\alpha > 0 \\
    \end{cases}\\
    \hat\sigma_y^2 &= \frac{1}{nq}\|Y - XL\beta\|_F^2.
\end{align}

\subsection{Kernel Supervised Dimension Reduction}
In this section we extend all proposed methods to perform kernel SDR.
\subsubsection{Kernel PCA}
Kernel PCA (kPCA) \cite{scholkopf1997kernel} is a means of performing non-linear unsupervised dimension reduction by performing PCA in a high-dimensional feature space associated to a symmetric positive definite kernel. 

Let $k: \R^p \times \R^p \to \R$ be a symmetric positive definite kernel function. Associated to $k$ is a high dimensional feature space $\mathcal{F}$ and mapping $\Phi$ such that $\Phi: \mathbb{R}^p \to \mathcal{F}$. The kernel matrix associated to $k$ is $K_{ij} = k(\b x_i,\b x_j)$, and $k(\b y,\b z) = \langle \Phi(\b y), \Phi(\b z) \rangle_\mathcal{F}$ $\forall \b y, \b z \in \R^p$. Let $X_\Phi$ be the matrix with $n$ rows where each row is the representation in $\mathcal{F}$ of the corresponding row of $X$. Note that kPCA finds the projection of $X_\Phi$ onto its top $r$ principal components, rather than the principal components themselves. Computing the principal components themselves is usually impractical or intractable as $\Phi$ may be unknown and/or $\mathcal{F}$ may be of arbitrarily high dimension. Computationally, all that is required is to find the eigenvectors corresponding to the $r$ largest eigenvalues of the centered kernel matrix $\widetilde K = K - \frac{1}{n}\b 1 \b 1'K - \frac{1}{n} K \b 1 \b 1' + \frac{1}{n^2}\b 1 \b 1' K \b 1 \b 1'$. This amounts to solving
\begin{align}
   \hat L = &\min_{L} \|\widetilde K - \widetilde KLL'\|_F^2 \label{kPCA} \\
   & s.t. \ \  L'L=I_r, \nonumber
\end{align}
where now $p=n$, and so $L$ is $n \times r$. Let $\{\b v_i\}_{i=1}^r$ be the top $r$ principal components in the new feature space $\F$, and let $V  = [ \b v_1, \b v_2, \dots, \b v_r]$. The columns of $\hat L$ are such that
\begin{equation*}
    \b v_i = \sum_{j=1}^n \hat L_{ji} \widetilde \Phi(\b x_j),
\end{equation*}
where $\widetilde \Phi(\b x_j) = \Phi(\b x_j) - \frac{1}{n}\sum_{j'=1}^n \Phi(\b x_{j'})$ is the centered representation of $\b x_j$ in $\F$. To ensure the $\b v_i$ are unit norm, the columns of $\hat L$ must be normalized to obtain $\bar L$ such that $\bar L' K\bar L = I_r$. The projection of a data point $\b x$ onto the $i^{th}$ component is
\begin{equation*}
    \langle \b v_i, \widetilde \Phi(\b x) \rangle_\mathcal{F} = \sum_{j=1}^n \bar L_{ji} \widetilde k(\b x,\b x_j),
\end{equation*}
where 
\begin{align*}
\widetilde k(\b y,\b z) &= \langle \widetilde \Phi(\b y), \widetilde \Phi(\b z)
\rangle_\mathcal{F} \\
&= k(\b y,\b z) - \frac{1}{n}\sum_{i=1}^n \left(k(\b y, \b x_i) + k(\b x_i, \b z) \right) \\
&+ \frac{1}{n^2}\sum_{j=1}^n\sum_{j'=1}^n k(\b x_j, \b x_{j'}).
\end{align*}
Most importantly for our purposes, the weights of the projection of the training data are given by
\begin{equation}
     \Pi_{\{\b v_i\}_{i=1}^r}(X)V = \widetilde K\bar L \in \R^{n \times r}, \label{kPCA_embedding}
\end{equation}
i.e., $(\widetilde K\bar L)_{ji} = \langle \b v_i, \widetilde \Phi(\b x_j) \rangle_\mathcal{F}$. We refer to using kPCA in procedures analogous to PCR and PCC as kPCR and kPCC, respectively.
\subsubsection{Kernel LSPCA and LRPCA}
We highlight the fact that $L$ does not have the same interpretation in the kernel setting as $L$ in the linear setting. As in kPCA, in the problems to follow $L$ provides coefficients for a low dimensional embedding and does not have a direct interpretation in terms of the importance of various features of the original data.

Recall that the projection of the training data is given by $\widetilde K\bar L \in \mathbb{R}^{n\times k}$. This suggests we could kernelize the proposed methods by substituting $\widetilde K$ for $X$. The problem is that the columns of $\bar L$ do not, in general, have unit norm in kPCA. Since the columns of $\bar L$ are just scaled versions of the columns of $\hat L$, there exists a $\bar \beta$ with scaled rows of $\beta$ such that 
\begin{equation*}
    \widetilde K\bar L \beta = \widetilde K \hat L \bar\beta,
    \vspace{-5pt}
\end{equation*}
where the columns of $\hat L$ have unit norm. Therefore we can just substitute the kernel matrix $\widetilde K$ for the data matrix $X$ in LSPCA and LRPCA, allowing the scaling to be absorbed by $\beta$. Similar to before, we can write the general kernel SPCA problem
\begin{align*} \label{kLSPCA}
&\min_{L, \beta, \lambda, \gamma} \ \ G(L, \beta, \lambda, \gamma; \widetilde K, Y)
\\  &s.t. \ \  L'L = I_r. 
\vspace{-10pt}
\end{align*}

\subsection{Manifold Conjugate Gradient Descent Algorithm}
Below, we state the manifold conjugate gradient descent algorithm \cite{edelman1998geometry}, specifically for the Grassmann manifold.

\begin{algorithm}
\caption{Manifold Conjugate Gradient Descent \cite{edelman1998geometry}}\label{alg:conjgrad}
\textbf{Input:} A cost function $G(L)$, a $p \times r$ orthogonal matrix $L_0$ \\
\textbf{Output:} A solution $L^*$
\begin{algorithmic}[1]
\Procedure{$\operatorname{MCGD}$}{$G(L), L_0$}
\State{$\Delta_0 \gets \operatorname{grad} G\big|_{L=L_0}$}
\State{$C_0 \gets - \Delta_0$}
\State $k \gets 0$
\Repeat
\item[\quad\quad\quad \ \Comment{Form compact SVD}]
\State $U\Sigma V' \gets \operatorname{svd}(C_{k})$ 
\item[\quad\quad\quad \ \Comment{Perform a line search}]
\State $t_k \gets {\min}_t G(L_{k}V\cos(\Sigma t)V' + U\sin(\Sigma t)V')$
\item[\quad\quad\quad \ \Comment{Update $L$ and search direction}]
\State $L_{k+1} \gets L_{k}V\cos(\Sigma t_k)V' + U\sin(\Sigma t_k)V'$
\State $\Delta_{k+1} \gets -(I_p - L_{k+1}L_{k+1}')(\frac{\partial G}{\partial L}\big|_{L=L_{k+1}})$
\State $\widetilde C_{k+1} \gets (-L_{k}V\sin(\Sigma t_k) + U\cos(\Sigma t_k))\Sigma V'$
\State{$A_k \gets L_{k}V\sin(\Sigma t_k)$}
\State{$B_k \gets U(I - \cos(\Sigma t_k))$}
\State $\widetilde \Delta_{k} \gets \Delta_{k} - (A_k + B_k)U'\Delta_{k}$
\State $d_k \gets \frac{\langle \Delta_{k+1} - \widetilde \Delta_{k}, \Delta_k \rangle}{\langle \Delta_{k}, \Delta_{k} \rangle}$
\State $C_{k+1} \gets -\Delta_{k+1} + d_k \widetilde C_k$
\If{$k \equiv 0 \ \operatorname{mod} \ r(p-r)$} 
\State $C_{k+1} \gets -\Delta_{k+1}$
\EndIf
\State $k \gets k+1$
\Until{Convergence}\\
\Return $L_k$
\EndProcedure
\end{algorithmic}
\end{algorithm}


\subsection{Links to Datasets}
The datasets used in this work that are directly available online are Ionosphere\footnote{Ionosphere: \url{https://archive.ics.uci.edu/ml/datasets/Ionosphere}\hfil}, Sonar\footnote{Sonar: \url{https://archive.ics.uci.edu/ml/datasets/Connectionist+Bench+\%28Sonar\%2C+Mines+vs.+Rocks\%29} \hfil}, Colon\footnote{Colon:\url{https://jundongl.github.io/scikit-feature/datasets.html} \hfil}, Arcene\footnote{Arcene:\url{https://jundongl.github.io/scikit-feature/datasets.html} \hfil}, Residential\footnote{Residential:\url{https://archive.ics.uci.edu/ml/datasets/Residential+Building+Data+Set} \hfil}, and Music\footnote{Music:\url{https://archive.ics.uci.edu/ml/datasets/Geographical+Original+of+Music} \hfil}. \hfil \vfill
\end{document}